\documentclass{article}

 \usepackage[preprint]{neurips_2026}

\usepackage[utf8]{inputenc} 
\usepackage[T1]{fontenc}    
\usepackage{hyperref}       
\usepackage{url}            
\usepackage{booktabs}       
\usepackage{amsfonts}       
\usepackage{nicefrac}       
\usepackage{microtype}      
\usepackage{xcolor}         

\usepackage{xspace} 
\usepackage{graphicx} 
\usepackage{fontawesome5}
\usepackage{amsmath}
\usepackage{cleveref}

\newcommand{\linkicon}[1]{\raisebox{-0.2ex}{#1}\,}

\newcommand{\hficon}{\linkicon{\includegraphics[height=1.05em]{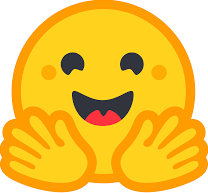}}}
\newcommand{\iconbox}[2][1.6em]{\makebox[#1][c]{#2}}

\usepackage{amssymb}
\usepackage{pifont}
\newcommand{\cmark}{\ding{51}} 
\newcommand{\xmark}{\ding{55}} 
\usepackage{adjustbox}
\usepackage[table]{xcolor}
\usepackage{makecell}
\usepackage{enumitem}

\usepackage[dvipsnames]{xcolor}
\newcommand{\greencheck}{{\color{ForestGreen}\cmark}}
\newcommand{\redcross}{{\color{red}\xmark}}
\usepackage{tabularx}
\usepackage{anyfontsize}   
\newcommand{\mediumpfont}{%
  \fontsize{7.5}{9}\selectfont
}

\usepackage{titlesec}
\titlespacing*{\section}      {0pt}{1ex plus 0.3ex minus 0.3ex}{1ex plus 0.2ex}
\titlespacing*{\subsection}   {0pt}{0.8ex plus 0.2ex minus 0.2ex}{0.6ex plus 0.1ex}
\titlespacing*{\subsubsection}{0pt}{0.6ex plus 0.2ex minus 0.2ex}{0.4ex plus 0.1ex}
\titlespacing*{\paragraph}    {0pt}{0.2ex plus 0.1ex minus 0.1ex}{1em}

\usepackage{multirow} 

\newcommand{\dataset}{CalTennis\xspace}

\title{\dataset: A Multi-View, Long-Range 3D Human Pose Estimation Benchmark of Skilled Tennis Motion}

\title{\dataset: A Multi-View Human Pose Benchmark of Tennis Videos}
\title{\dataset: Human Pose Estimation Benchmark from Multiview Tennis Video}
\title{\dataset: Large Multi-View Tennis Video Dataset and Benchmark of Monocular-to-3D Pose Estimation}

\author{%
    Ilona Demler \quad
  Xinran Xie\quad
  Blake Werner\quad
  Anna Szczuka\quad
  Pietro Perona \\
  California Institute of Technology \\
  \texttt{\{idemler, perona\}@caltech.edu}
}
\begin{document}

\maketitle

\begin{abstract}

The Caltech Tennis Dataset (\dataset) is a large-scale video benchmark for evaluating monocular-to-3D pose estimation in the wild. \dataset comprises over 11 million frames (51 hours) of tennis practice and match play from 40 players, captured with 2–6 synchronized cameras at 60Hz. It is 10× larger than existing in-the-wild human motion video datasets and 3× larger than existing MOCAP-ground-truthed datasets, and it is the first large-scale benchmark to provide synchronized multi-view recordings of expert athletic motion. The multi-view setup enables inexpensive, label-free evaluation of monocular-to-3D pose estimation algorithms. We describe a simple, standardized protocol that enables data collection without specialized equipment or expertise, along with fully automated video calibration and synchronization. Benchmarking state-of-the-art monocular-to-3D pose methods on \dataset, we find that while 3D joint angle recovery is now quite accurate, all models struggle to estimate depth and foot contact consistently. We further propose two novel performance metrics -- footwork and stability -- as well as qualitatively study body shape inconsistency. These metrics expose previously underexplored failure modes and point to concrete opportunities for improvement in pose estimation and action analysis. 

   \textbf{\iconbox{\faGlobe}~Project Page:} \href{https://ilonadem.github.io/caltennis-website/}{https://ilonadem.github.io/caltennis-website/} \\
  \textbf{\iconbox{\hficon}~Dataset:}
  \href{https://huggingface.co/datasets/demalenk/caltennis}{https://huggingface.co/datasets/demalenk/caltennis}
  
\end{abstract}

\section{Introduction}


Estimating three-dimensional human motion from video is a critical task in a number of domains. 
Accurate pose measurement underpins applications from gait analysis and injury rehabilitation in healthcare~\cite{colyer2018review, wade2022markerless}, coaching and analytics in sports~\cite{thomas2017cvsports, uhlrich2023opencap}, character animation for movies and gaming~\cite{menache2011mocap, lombardi2018deepappearance}, and even pedestrian safety~\cite{rasouli2020autonomous} and gait identification in forensic analysis~\cite{sepasmoghaddam2023deepgait}.
More recently, human motion data has become central to robotics and embodied AI, where it is used for imitation learning, teleoperation, and training humanoid policies from human demonstrations ~\cite{he2024omnih2o}.
Across all of these, the utility of pose estimation is bounded by its accuracy: small errors in joint position, depth, or ground contact propagate directly into the biomechanical, behavioral, and narrative
conclusions that downstream systems draw.
Modern MOCAP remains the gold standard for accuracy, and it is the source of ground truth for the majority of current pose estimation benchmarks. But MOCAP is expensive, costing upwards of \$150{,}000 per installation~\cite{uhlrich2023opencap}, requiring a dedicated laboratory space, and constraining natural movement through body-worn markers and suits, making it impractical to deploy in the wild. 
Replacing MOCAP with monocular video analysis from a single ordinary camera, such as the phone in a coach's pocket or a spectator's hand, would therefore have a transformative effect in applications from medicine to sports and entertainment. While monocular video-based methods still fall short of MOCAP-grade accuracy, the last decade has seen considerable progress toward this goal~\cite{bogo2016smplify, kanazawa2018hmr, kolotouros2019spin, kocabas2020vibe, goel2023humans, shin2024wham}.  Large and challenging benchmark datasets are now needed to highlight the remaining failure modes and guide researchers towards reaching application-grade accuracy. 


We identify
six criteria for the benchmark needed to guide this progress:
\begin{enumerate}[itemsep=0pt, topsep=0pt, leftmargin=*]
    \item {\bf In-the-wild}: Record people in real-life, unconstrained, unscripted, everyday activities.
    \item {\bf Pose coverage}: Cover body poses beyond the limited space of walking or other routine activity.
    \item {\bf Meaningful, repeated actions}: Record meaningful actions, repeated by different actors, to support research on higher-level behavior and activity analysis.
    \item {\bf Expert relevance}: Capture activities relevant to users (e.g coaches and clinicians), whose expertise can guide the field to the fine-grained accuracy that applications demand.
    \item {\bf Large-scale}: Large enough to support training and evaluation at modern scale, and to contain the rare corner cases on which models most often fail.
    \item {\bf Easy and inexpensive to collect}: Researchers should be able to contribute equivalent datasets to broaden diversity, and existing ones should be expandable as needed. It should be possible for anyone to collect the data with minimal instruction, so that the collection can scale easily.
\end{enumerate}
Methodologically, the last criterion suggests {\it evaluation from the data itself}, without prohibitive ground truth such as MOCAP, body sensors, or human annotations used by all current benchmarks.

\begin{figure}[t]
  \centering
  \includegraphics[width=0.9\linewidth]{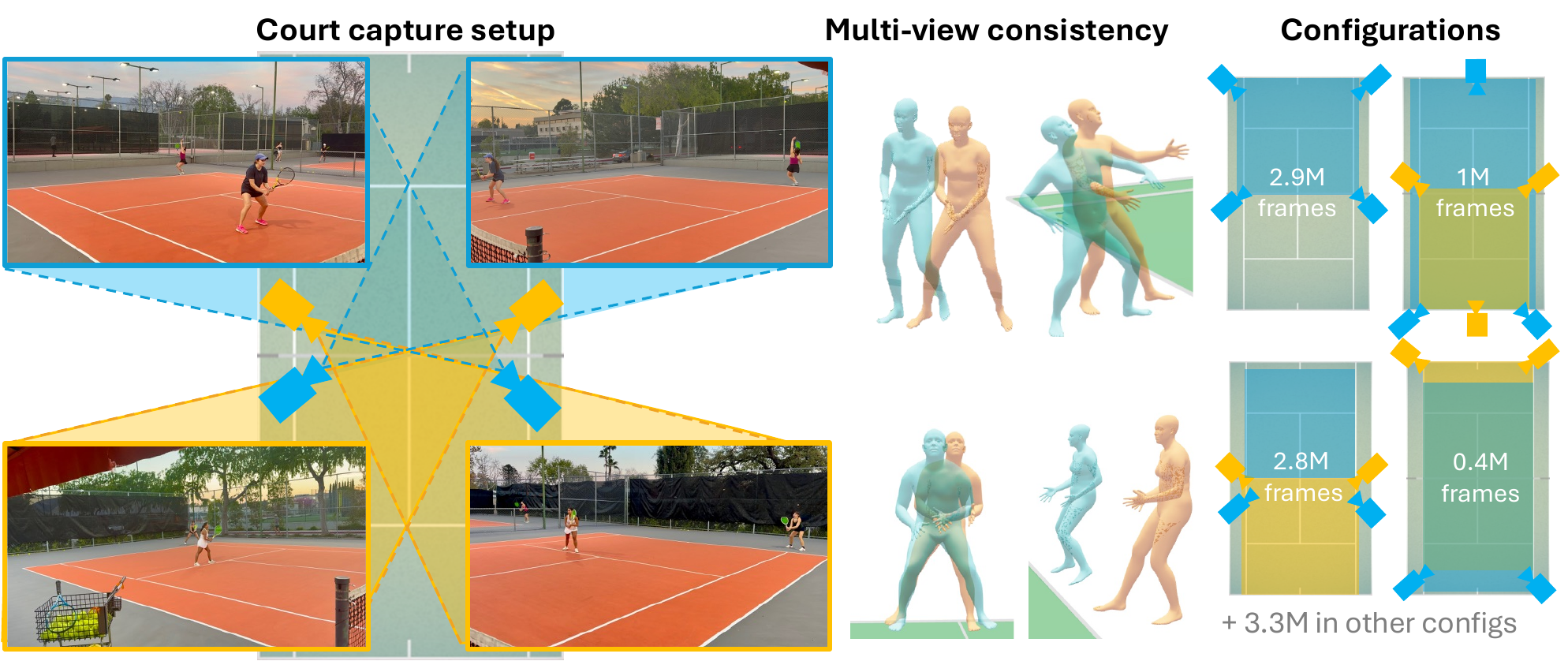}
  \caption{\textbf{Overview of \dataset Setup}. 
  {\bf (Left)}: 4-tripod setup, with two overlapping camera views (blue \& orange) on each half-court (\cref{sec:cam_calib,sec:data-collection}) {\bf (Center)}: Overlapping views enable multi-view consistency evaluation. We measure the difference in 3D position, the difference in pose once the difference in 3D position is removed, as well as body shape and foot contact (\cref{sec:metrics}). {\bf (Right)}: We collect up to 6 concurrent views of gameplay across varying configurations, testing reconstruction consistency against diverse view and depth distributions (\cref{sec:dataset}).}
  

  \label{fig:dataset-overview}
\end{figure}

Sports are a compelling source of data: motion is fast and diverse, actions are meaningful and repeated, and data is available across skill levels --- ideal for studying technique, style, proficiency, and model consistency. Despite this natural fit, sports remain underrepresented in motion analysis benchmarks for computer vision algorithms. Furthermore, current models generalize poorly to athletic motion~\cite{yeung2025athletepose3d}, distorting downstream biomechanical conclusions~\cite{camomilla2017jointerrorfactors}. Amongst sports, tennis is particularly attractive: standardized court markings let camera geometry be recovered automatically from any view, it spans a rich action vocabulary (serves, volleys, sprints), players are rarely occluded, it may be analyzed at a fine spatio-temporal scale (individual shots, footwork), at an intermediate one (court position, ball placement) and at a long time-scale (tactics, strategy, coaching advice). Because tennis is played worldwide, equivalent data can be collected by independent teams at low cost, satisfying the practical requirements above.

We introduce the Caltech Tennis Dataset (\dataset), the first large-scale, multi-view video dataset of real-world tennis practice and match play. \dataset contains over 11 million frames (51 hours) from 40 players, captured with 2--6 synchronized consumer cameras at 60Hz. It is 10× larger than prior in-the-wild benchmarks, with substantially more variety in depth from camera, people per video, and pose coverage. To our knowledge, it is the only large-scale multi-view video dataset of unscripted human motion in natural environments. Overlapping views let us evaluate monocular pose estimators without privileged ground truth: a correct reconstruction must agree across views, and inter-view disagreement lower-bounds each model's error. This lets us probe errors in pose estimates, specifically foot contact, body shape, depth, and stability, that are invisible to existing benchmarks.

We benchmark five state-of-the-art monocular 3D pose estimators and find that on \dataset they are overall significantly less accurate than on previous benchmarks. More in detail: we find that joint-angle pose recovery is often accurate, while metric-scale depth estimates are highly unstable, producing rapid, unrealistic jumps in estimated body position. Furthermore foot-contact detection is inconsistent across frames and views. Body shape estimation is also inconsistent: limb lengths, height, and body proportions of the same person vary across views, and differ systematically across models. These are quantities that balance, force, and speed estimation in biomechanical analysis depend on.

This work makes the following contributions:
\begin{enumerate}[itemsep=0pt, topsep=0pt, leftmargin=*]
\item \textbf{Dataset.} \dataset, the largest multi-view, real-world video dataset of skilled athletic motion with 11+ million frames of tennis practice and match play of 40 players captured with up to 6 synchronized consumer cameras.
\item \textbf{Evaluation methodology.} A label-free framework for evaluating monocular 3D pose estimators using multi-view consistency as a lower bound on error, together with new inconsistency metrics: stability, foot skating, and body shape, that expose failure modes invisible to standard benchmarks.
\item \textbf{Benchmark of state-of-the-art monocular 3D pose algorithms.} An extensive evaluation of five state-of-the-art monocular 3D pose methods on \dataset, showing that while joint-angle pose is often accurate, depth, foot contact, and body shape estimates are unstable — quantities on which downstream applications depend.
\item \textbf{Data-collection recipe.} A simple, inexpensive protocol for capturing multi-view tennis video using consumer phones and lightweight collapsible tripods; calibrated via court geometry; and designed so that equivalent datasets can be easily collected by other teams.
\end{enumerate}

\section{Related work}

\paragraph{Human pose estimation methods.}

Reconstructing human poses from images and videos is typically formulated as estimating SMPL \cite{black2015smpl} or SMPL-X \cite{pavlakos2019smplx} parameters, and has evolved through several increasingly challenging paradigms.
Early methods targeted cropped single-person images \cite{tao2022vitpose, kanazawa2018hmr, kanazawa2018endtoendrecovery, black2021pare, kolotouros2019spin, magnusson2023cliff}, then arbitrary images \cite{magnusson2023multihmr, daniildis2020coherentreconmultihumans, cai2024aios, kitani2026sam3dbody}, leveraging pretrained object detectors \cite{tao2022vitpose, redmon2016yolo, girshick2019detectron2} before predicting 3D poses in camera coordinates.
The problem was next extended to videos with a temporal processing step for trajectory continuity \cite{daniildis2020coherentreconmultihumans, kanazawa2019hmmr, black2021pare, hoogs2020meva}. Current video-based methods reconstruct human motion in global world coordinates \cite{iqbal2023pace, iqbal2024coin, shin2024wham, daniilidis2024tram, kanazawa2023slahmr, kautz2022glamr, wang2025prompthmr, yuan2025genmo, xiaowei2024gvhmr}. Because estimating depth is ill-posed, most approaches run SLAM or camera calibration steps and learn body priors.
Our work evaluates state-of-the-art 3D human pose reconstruction methods on videos to assess their utility for downstream applications.
While we use SMPL/SMPL-X in this evaluation, our method is agnostic to the representation. 

\paragraph{Benchmarks.}
Collecting ground-truth 3D human motion at scale is challenging and expensive.  Laboratory benchmarks such as  Human3.6M~\cite{sminchiescu2014human36m} record one person at a time in a $4\,\text{m} \times 3\,\text{m}$ area with synchronized cameras and time-of-flight sensors. In-the-wild benchmarks relax the environment but retain body-worn instrumentation: 3DPW~\cite{black20183dpw} attaches IMUs and EMDB~\cite{kaufmann2023emdb} uses wireless electromagnetic sensors on people moving through everyday outdoor spaces. Contact-focused benchmarks such as RICH~\cite{black2022rich} rely on laser-scanned scene geometry, while HI4D~\cite{yin2023hi4d} uses 4D scans of close human--human interaction to optimize meshes. None record unscripted, skilled motion in natural environments at scale, and additional views, when present, are used to \emph{generate} ground truth rather than evaluate it.
We discuss the resulting coverage gaps quantitatively in Section~\ref{sec:dataset}.

\paragraph{Multi-view capture and label-free evaluation.}
A parallel line of work uses multi-camera rigs to capture human motion without body-worn markers. CMU Panoptic Studio~\cite{joo2015panoptic} uses a geodesic dome of 500+ synchronized cameras, and AIST++~\cite{li2021ai} provides multi-view dance recordings in a controlled studio. At the lightweight end, OpenCap~\cite{uhlrich2023opencap} shows that two smartphones with a markerless pipeline recover clinically useful biomechanics. Our capture recipe sits in the same lightweight regime but targets unscripted, high-speed athletic motion in public facilities rather than controlled clinical settings. Our use of multi-view information is different: we treat view disagreement not as a training signal or a ground-truth generator, but as a direct, label-free measurement of reconstruction error. 

\paragraph{Sports video and pose datasets.}
Sports video has a long history in computer vision, though most large-scale sports datasets such as Penn Action~\cite{zhang2013actemes}, FineGym~\cite{shao2020finegym}, SoccerNet~\cite{giancola2018soccernet}, and SportsMOT~\cite{cui2023sportsmot} support action recognition, temporal localization, and multi-object tracking, but do not provide 3D pose annotations. A smaller set of recent datasets does target 3D pose in sports, and these are CalTennis's closest relatives. Recent 3D pose datasets, such as AthletePose3D~\cite{yeung2025athletepose3d} and SportsPose~\cite{ingwersen2023sportspose} are restricted to lab-based, imitated motions. AthletePose3D additionally provides a small subset of multi-view ice rink jumping videos. These videos are similar to our closest relative, WorldPose~\cite{jiang2024worldpose}, which proves unscripted athletic motion can be captured at scale from multi-view video. \dataset shares this premise but differs from the two datasets in three key ways: (1) we use accessible consumer phones on tripods rather than elite broadcast infrastructure; (2) we leverage multi-view disagreement as a label-free error metric instead of triangulating pseudo-ground-truth; and (3) we focus on highly articulated strokes and footwork rather than large-scale locomotion. Thus, \dataset, WorldPose, and AthletePose3D serve as complementary benchmarks.

\section{The \dataset Dataset}
\label{sec:comparison_to_others}
\label{sec:dataset}

\newcolumntype{C}{>{\centering\arraybackslash}X}
\begin{table}[]
\label{tab:calten_vs_others}
\mediumpfont
\caption{\textbf{\dataset vs. Other Benchmarks.} Our videos span significantly longer frame durations from up to 6 concurrent views at a time. We capture more people per video, at greater depth ranges and with more pose coverage. We report numbers on the test splits of all datasets.}

\centering
\setlength{\tabcolsep}{3.5pt} 

\resizebox{\textwidth}{!}{
\begin{tabular}{lcccccccccccc @{}}

\toprule

 & \makecell{Multi-\\View?}
 & {\makecell{Real-\\world?}}
 & {\makecell{Num\\Frames (M)}} 
 & {\makecell{Avg. seq \\ len (sec)}}
 & {\makecell{Depth\\range (m)}}
 & {\makecell{Per-Joint \\ Articulation}}
 & {\makecell{Pose Space \\ Coverage}}
 & {\makecell{Collector\\Friendly?}}
 & {\makecell{Hardware\\Cost k\$}}
 & {\makecell{Accuracy}}
 \\

\midrule

\textbf{3DPW~\cite{black20183dpw}}         
    & \redcross 
    & \greencheck 
    & \phantom{0}0.05 
    & \phantom{0}\phantom{0}45 
    & \phantom{0}3.1 --\phantom{0}7.4
    & 69.9\%
    & 58\% 
    & \redcross 
    & \phantom{000.}21
    & mm \\

\textbf{EMDB~\cite{kaufmann2023emdb}}         
    & \redcross 
    & \greencheck 
    & \phantom{0}0.11 
    & \phantom{0}\phantom{0}42 
    & \phantom{0}1.9 --\phantom{0}2.7
    & 69.9\%
    & 60\% 
    & \redcross 
    & \phantom{000.}31 
    & cm \\

\textbf{RICH~\cite{black2022rich}}         
    & \greencheck 
    & \greencheck 
    & \phantom{0}0.54
    & \phantom{0}127 
    & \phantom{0}4.2 --\phantom{0}4.7
    & 75.5\%
    & 62\%
    & \redcross 
    & \phantom{00.}100
    & cm \\

\textbf{HI4D~\cite{yin2023hi4d}}         
    & \redcross 
    & \redcross 
    & \phantom{0}0.01
    &  \phantom{0}\phantom{0}\phantom{0}7 
    & \phantom{0}2.7 --\phantom{0}3.0
    & 71.7\%
    & 70\%  
    & \redcross 
    &  \phantom{000.}21 
    & mm \\

\textbf{Human3.6M~\cite{sminchiescu2014human36m}}    
    & \greencheck 
    & \redcross 
    & \phantom{0}1.47
    & \phantom{0}340 
    & \phantom{0}4.5 --\phantom{0}5.8
    & -
    &  89\%
    & \redcross 
    & \phantom{00.}150
    & mm \\

\textbf{SportsPose~\cite{ingwersen2023sportspose}}    
    & \greencheck 
    & \redcross 
    & \phantom{0}1.50
    & \phantom{000}9
    & \phantom{0}0.7 --\phantom{0}3.4
    & -
    & 47\%
    & \redcross 
    & \phantom{000.}25 
    & cm \\


\rowcolor{gray!15}
\textbf{\dataset}  
    & \greencheck 
    & \greencheck 
    & 11.03
    &  3365
    & 13.4 - 16.7 
    & 70.2\%
    & 85\% 
    & \greencheck 
    & \phantom{0000.}2 
    & cm \\
\bottomrule
\end{tabular}
}
\end{table}

\begin{figure}
  \centering
  \begin{minipage}{0.62\textwidth}
    \centering
    \includegraphics[width=\linewidth]{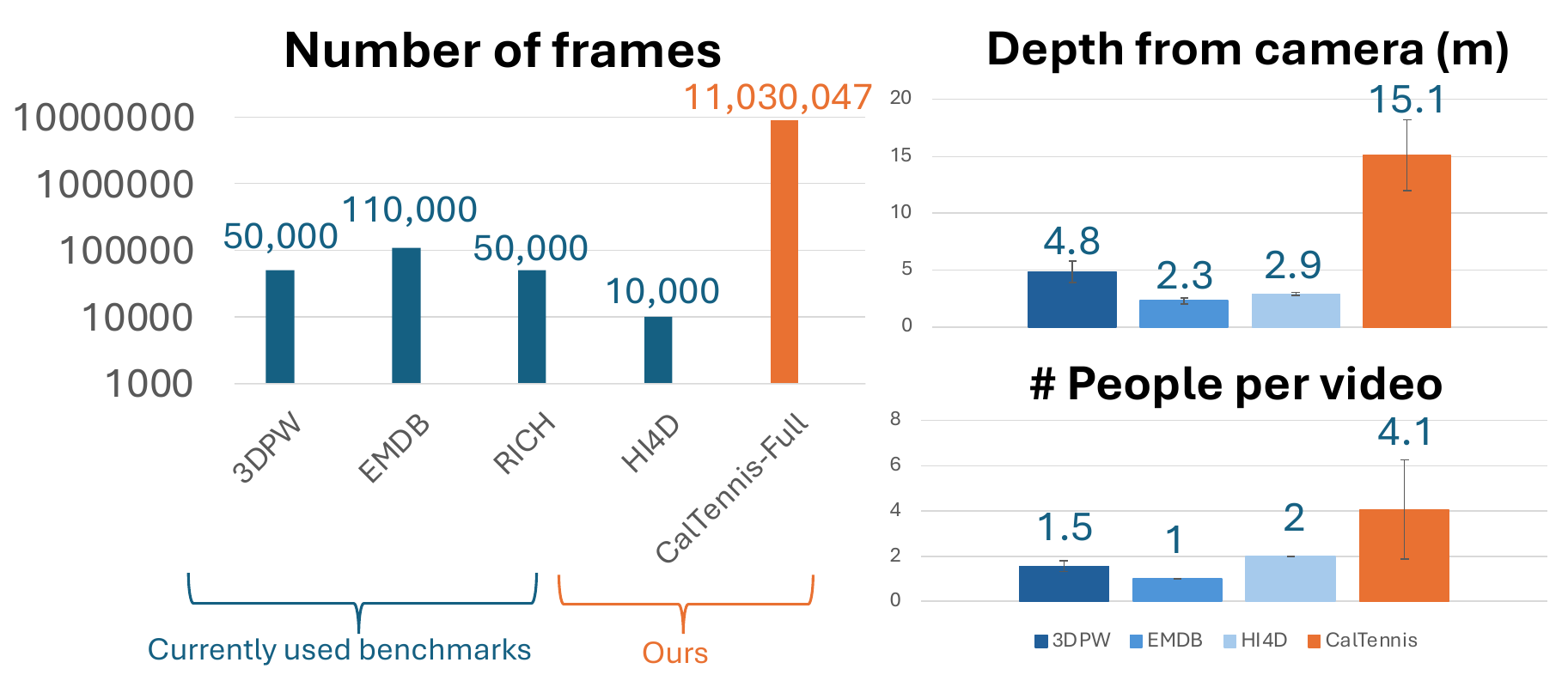}
  \end{minipage}
  \hfill
  \begin{minipage}{0.36\textwidth}
    \centering
    \includegraphics[width=\linewidth]{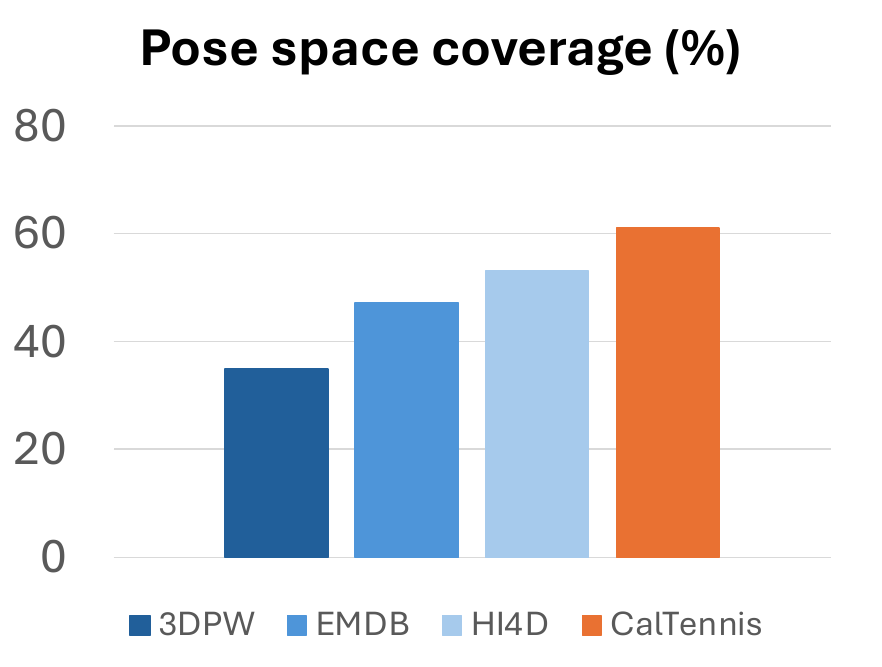}
  \end{minipage}
  \caption{\textbf{\dataset complexity compared to other real-world benchmarks.} {\bf(Left)}: \dataset contains 10× frames than currently used benchmarks. {\bf(Middle)}: \dataset contains more variation in the distance of people from the camera (top), as well as many more people per video (bottom). {\bf(Right)}: \dataset has the highest pose space coverage (defined in \S\ref{sec:comparison_to_others}). }
  \label{fig:plots_of_dataset_comparisons}
\end{figure}

\subsection{Data collection and curation}

\label{sec:data-collection}

Our \dataset benchmark contains real-world multi-view videos of tennis practices and games. We opted for a multi-view setup to enable  inexpensive, label-free evaluation of monocular-to-3D pose estimation algorithms. We designed a collection protocol to minimize  equipment cost and make it easy for anyone to collect good-quality video. In our case it is the players themselves collecting video of the practices and games of the Caltech men's and women's tennis teams with minimal instruction. Video is recorded at 60\,Hz in HD ($\!1920 \times 1080\!$ pixels) using the main camera (1x, 24mm f/1.78) of standard iPhones (model 14 or newer, owned by players and coaches at no marginal cost to the project) on fully extended 1.65\,m MagSafe tripods (costing $\! \approx\!$ $\!  $\! \$40 each) placed around the court at standard locations and orientations specified by us (\cref{fig:dataset-overview}). We deploy 2-6 cameras concurrently per session on standard courts (each half-court measuring 10.98\,m $\! \times\! $ 11.88\,m), spanning both half-court and full-court; cameras are typically $\! \approx\! $ 12m apart at 1.65\,m height (\cref{fig:dataset-overview}). Across sessions by the Caltech tennis teams and recreational players, we collected over 11 million frames (51 hours) from 40 players spanning collegiate to recreational levels, with 400K--2.9M frames per camera configuration, allowing us to evaluate multi-view reconstruction consistency across diverse view geometries.  All sessions were conducted under IRB approval with voluntary, informed consent, and we blur all faces using the deface algorithm~\cite{orbhd_deface} to protect player privacy.

\subsection{Dataset statistics}
\dataset is the first benchmark to use multi-view, real-world recordings of skilled human motion, capturing data underrepresented in existing pose datasets and more representative of downstream motion-reconstruction applications. \dataset contains $10\times$ more frames than other real-world benchmarks and $3\times$ more than the largest MoCap benchmark (Human3.6M), with substantially greater depth variability: 90\% of poses lie 13.4--16.7m from the camera, versus 4.5--5.8m in Human3.6M. We measure pose diversity along two axes. Pose space coverage is the Shannon entropy of frame-to-cluster assignments (over $k=500$ PCA clusters of the shared pose-joint space), normalized so 100\% indicates uniform coverage. Per-joint articulation is the entropy of each joint's angular distribution divided by its anatomical range of motion, averaged across joints. \dataset scores 85\% on coverage, versus 62\% for the next-best real-world benchmark and 89\% for lab-captured Human3.6M. Its per-joint articulation (0.70) is comparable to other real-world benchmarks (0.67--0.76), since tennis is an upright sport that constrains spine and pelvis motion; on the joints the sport actively recruits — knees, shoulders, and elbows — \dataset has the highest articulation of any benchmark (per-joint histograms in Appendix~\ref{app:further_complexity_anlyses}). High coverage with sport-specific articulation indicates a dense, domain-specific motion manifold.

\section{Calibration and Synchronization}

To run multi-view label-free evaluations of SOTA human pose estimation models, we lift pose predictions from each camera into a shared global space-time reference frame by calibrating cameras and synchronizing the videos. We review our method below, and more fully in Appendix ~\ref{app:camera_calib}.

\begin{figure}[]
  \centering
  \includegraphics[height=4cm]{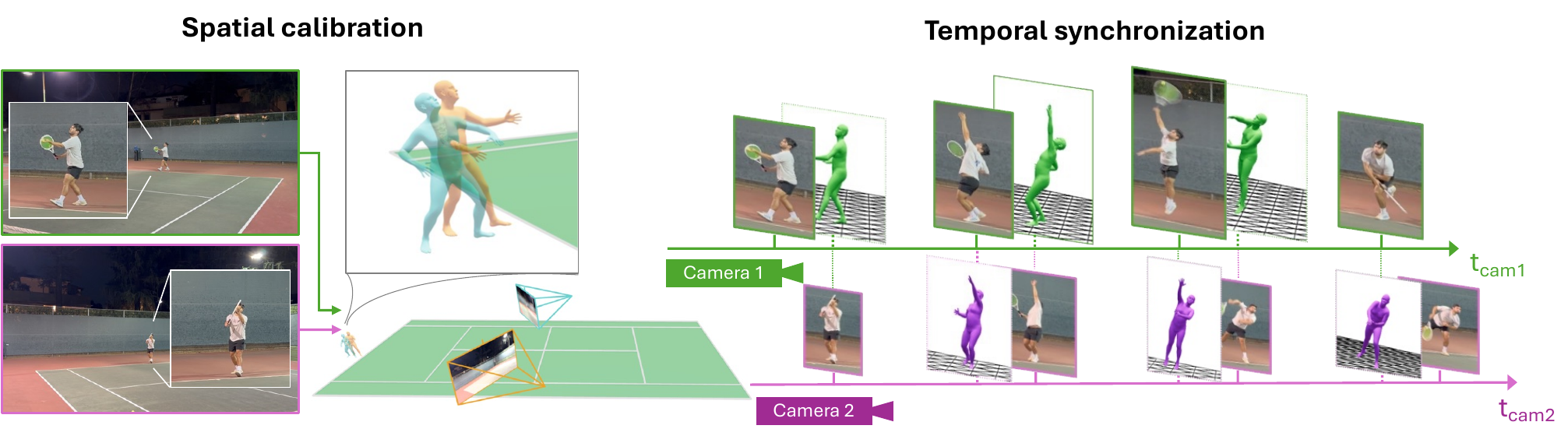}
  \caption{\textbf{Spatiotemporal calibration and synchronization.}
  {\bf(Left)}: Calibrating cameras (intrinsic and extrinsic calibration, outlined in \cref{sec:cam_calib}) allows us to lift model estimates into a shared court coordinate system (\S\ref{sec:spat_fusion}). Discrepancy in depth estimates results in differing 3D translation estimates. {\bf(Right)}: Videos lack identical timestamps, so we align sequences by optimizing a continuous global offset variable $\Delta t$ (\S\ref{sec:temp_calib}) to minimize pose misalignment.
  }
  \label{fig:time-calib}
\end{figure}

\paragraph{Problem definition.}
Given $N$ cameras facing a scene, each camera $c^i$ records a video $V^i = \smash{\{I_t^i\}_{t=0}^{t_k^i}} \in \mathbb{R}^{H \times W \times 3}$ with timestamps $t \in \{ t^i_0, …, t^i_k \} \in \mathbb{R}^{+} $ and intrinsics $K^i \in \mathbb{R}^{3 \times 4}$ and extrinsics $(R^i, T^i) \in SO(3) \times \mathbb{R}^3$. We parameterize the poses of $p$ people via SMPL-X~\cite{pavlakos2019smplx}: $H^i = \smash{\{(\tau^i_t, \phi^i_t, \beta^i_t, \theta^i_t)\}_{t=t_0^i}^{t_k^i}}$, with translation $\tau^i_t \in \mathbb{R}^{p \times 3}$, body pose $\theta^i_t \in \mathbb{R}^{p \times 21 \times 3}$, orientation $\phi^i_t \in \mathbb{R}^{p \times 3}$, and shape $\beta^i_t \in \mathbb{R}^{p \times 10}$. Given per-view pose estimates $H^i = M(V^i)$ from a pre-trained model $M$, we lift $\{H^0, \dots, H^N\}$ into a shared spatio-temporal world frame to evaluate multi-view consistency (\cref{fig:time-calib}). Because timestamps are unsynchronized and extrinsics are unknown a priori, the following sections describe our camera calibration, spatial fusion, and temporal calibration steps.

\paragraph{Video recording setup and camera calibration.}
\label{sec:cam_calib}
We position up to $N=6$ cameras around the court to cover all sides of the player (Fig.\ref{fig:dataset-overview}). To calibrate each $c_i$, we exploit the standardized court geometry: in a frame $I_t^i$, we identify $n$ court line intersections $\{\hat{P}^i_k\}_{k=1}^n \in \mathbb{R}^3$ and their pixel correspondences $\{\hat{p}^i_k\}_{k=1}^n \in \mathbb{R}^2$. With intrinsics $K^i$ from iPhone metadata and projection $\pi(\cdot; K^i): \mathbb{R}^3 \to \mathbb{R}^2$, we minimize reprojection error to get extrinsics:
\begin{equation}
    \min_{R^i, T^i} \sum_{k=1}^{n} \left\| \pi(R^i {\hat{P}}_k + T^i; K^i) - {\hat{p}}_k \right\|^2
\end{equation}

\paragraph{Spatial fusion.}
\label{sec:spat_fusion}
With camera extrinsics $(R^i, T^i)$ recovered, we lift estimates into the world frame using the model-to-world transform $T_{W_i^{\text{model}} \to W}$, which maps each translation as $\tilde{\tau}_t^i = T_{W_i^{\text{model}} \to W}\, [\tau_t^i; 1]$. We provide further details in Appendix \ref{app:camera_calib}.

\paragraph{Temporal synchronization.}
\label{sec:temp_calib}
Athletic motion requires millisecond precision. iPhone cameras receive global timestamps up to the nearest second, resulting in up to 1000-millisecond offsets between videos; ignoring this lack of alignment and comparing nearest-frame predictions from two videos would conflate model error with temporal misalignment. We resolve this in two stages. First, we linearly interpolate reconstructed poses to query any camera $c^j$ at an arbitrary time $t$. Second, to fix up to 1000~ms of inter-camera offset caused by coarse device logging, we optimize a global offset $\Delta t \in [-1000, 1000]$~ms via grid search to minimize cross-view pose disagreement (\cref{fig:time-calib}).

\section{Evaluation Metrics}
\label{sec:metrics}


\dataset uses multi-view video recordings for label-free evaluation of monocular pose estimates: a correct prediction must agree across views, and inter-view disagreement lower-bounds each model's error. In addition to the standard metrics of MPJPE, PA-MPJPE, and PVE~\cite{kanazawa2018endtoendrecovery}, we introduce further metrics that capture additional notions of correctness that are relevant for downstream applications.

\paragraph{Translation error.} The L2 distance between per-view ($i,j$) translation estimates for person $p$:
\begin{equation}
    E_\text{trans} = \frac{1}{TP} \sum_{t, p} \| \tau_{i,p}^{t} - \tau_{j,p}^t \|_2.
\end{equation}

\paragraph{Pose error.} Mean per-joint position error relative to the pelvis coordinate. This measures orientation and pose consistency independently of translation error.  $J_i, J_j$ are joint estimates recovered from the per-view SMPL-X poses:
\begin{equation}
    E_{\text{pose}} = \frac{1}{T P K} \sum_{t, p} \sum_{k=1}^K \| J_{i,p,k}^{t} - J_{j,p,k}^t \|_2.
\end{equation}

\paragraph{Footwork.} 
We measure cross-view agreement of foot joint velocities $\mathbf{v}^{(i)}_{p,k,t} \in \mathbb{R}^3$ as well as foot heights $h^{(i)}_{p,k,t} \in \mathbb{R}$ for each view $i$ (e.g. foot floating), exposing footwork failures:
\begin{equation}
    E_\text{skate} = \frac{1}{Z} \sum_{t,p,k} \left\| \mathbf{v}^{(i)}_{p,k,t} - \mathbf{v}^{(j)}_{p,k,t} \right\|_2, \quad
    E_\text{height} = \frac{1}{Z} \sum_{t,p,k} \left| h^{(i)}_{p,k,t} - h^{(j)}_{p,k,t} \right|,
\end{equation}
where $Z = T P K_\text{foot}$ normalizes across frames, persons, and foot joints $k$.
 
\paragraph{Stability.} Pose stability determines whether motions are balanced, and models that disagree on stability across views are likely unreliable for sport-analysis. Following robotics literature ~\cite{mcghee_stability_1968}, we define per-view stability as the L2 distance from the projected center of mass to the convex hull $Q$ of grounded foot joints, set to zero when the projection lies inside $Q$:
\begin{equation}
    E^{(i)}_{\text{stab}} = \begin{cases}
    \min_{q \in Q} \| \mathrm{CoM}_{xy} - q \|_2 & \text{if } \mathrm{CoM}_{xy} \notin Q \\
    0 & \text{otherwise.}
    \end{cases}
\end{equation}
Cross-view stability error is the mean absolute disagreement between matched persons across views; large values indicate models disagree on whether a pose is balanced:
\begin{equation}
    E_{\text{stab}} = \frac{1}{TP} \sum_{t,p} \left| E_{\text{stab},p,t}^{(i)} - E_{\text{stab},p,t}^{(j)} \right|.
\end{equation}

\section{Experimental Evaluation}
We assess state-of-the-art human pose reconstruction models on \dataset and identify overlooked challenges in pose reconstruction that make it a difficult problem. 

\paragraph{Baselines.}

We evaluate five state-of-the-art architectures that report top performance on EMDB~\cite{kaufmann2023emdb}, RICH~\cite{black2022rich}, and 3DPW~\cite{black20183dpw} (Appendix~\ref{app:sota_model_perf}), spanning a range of strategies: TRAM~\cite{daniilidis2024tram} predicts per-person SMPL-X poses and lifts to world coordinates; GVHMR~\cite{xiaowei2024gvhmr} uses intermediate gravity-view coordinates; GENMO~\cite{yuan2025genmo} is a video-conditioned diffusion model; WHAM~\cite{shin2024wham} refines ground-foot contacts; and PromptHMR~\cite{wang2025prompthmr} conditions a transformer on prompts such as 2D keypoints and contacts. We report multi-view consistency on the standard metrics (Translation Error, Pose Error, MPJPE, PA-MPJPE) alongside Foot-Skating and Stability. We do not supply camera coordinates a priori, so each model runs its own camera-estimation preprocessing. We lift outputs into a shared coordinate system following Section~\ref{sec:cam_calib}. Experiments are run on an NVIDIA H100, on the first 5M frames of the dataset.

\subsection{Model Performance on \dataset}
\label{sec:overall_performance}
\newcolumntype{C}{>{\centering\arraybackslash}X}
\begin{table}[t]
\mediumpfont
\caption{
\textbf{Overall Model Performance.} We report multi-view consistency of poses estimated by SOTA models run on \dataset (\S\ref{sec:overall_performance}). Results are in millimeters ($mm$), except for foot-velocity ($m/s$). We define these metrics in \S\ref{sec:metrics}. Different models excel at different aspects of motion reconstruction: PromptHMR produces the most consistent translation and pose estimates, whereas WHAM produces the most consistent foot velocity estimates. All metrics computed on the first 5M frames of \dataset; full results forthcoming.
}
\label{tab:pose-agreement}
\begin{tabular*}{\textwidth}{@{\extracolsep{\fill}} l c c c c c c c c @{}}
\toprule
 & \textbf{Translation} & \textbf{Pose} & \textbf{MPJPE} & \textbf{PA-MPJPE} & \textbf{Foot-Vel} & \textbf{Foot-Height} & \textbf{Stability} \\ 
\midrule

\textbf{PromptHMR~\cite{wang2025prompthmr}} 
    & \textbf{\phantom{0}942} 
    & \textbf{105} 
    & 1,785 
    & \textbf{\phantom{0}84 }
    & 3.23 
    & \phantom{0}70 
    & 25 \\

\textbf{WHAM~\cite{shin2024wham}}      
    & 2,664 
    & 106
    & 2,675 
    & 119
    & \textbf{0.72}
    & 150
    & 44 \\

\textbf{GVHMR~\cite{xiaowei2024gvhmr}} 
    & 3,587
    & 109 
    & 1,066 
    & \phantom{0}88
    & 2.49
    & \textbf{\phantom{0}60 }
    & 21 \\

\textbf{TRAM~\cite{daniilidis2024tram}}      
    & 2,340 
    & 115
    & \phantom{0}958 
    & \phantom{0}91 
    & 6.65
    & \phantom{0}80 
    & 33 \\

\textbf{GENMO~\cite{yuan2025genmo}}      
    & 2,560 
    & 110 
    & 1,020
    & \phantom{0}91
    & 4.40
    & \phantom{0}60
    & \textbf{16} \\

\hline
\end{tabular*}
\end{table}

We evaluate the consistency of multi-view estimates made by state-of-the-art monocular 3D human pose reconstruction models, and report model performance in Table \ref{tab:pose-agreement}. We find that different models excel at different aspects of performance. PromptHMR performs best on the standard pose metrics -- translation error, pose error, and PA-MPJPE -- with an average of 0.942m, .105m, and .084m respectively. WHAM performs significantly worse on the standard metrics (2.664m translation error and 11.9cm PA-MPJPE), but performs exceptionally well on foot velocity consistency (.72m/s), compared to 3.23m/s for PromptHMR. We hypothesize that this is due to its iterative pose refinement step that optimizes foot-ground contacts.  GENMO poses are the most consistent along the foot height (.06m) and stability metrics. Overall, the highest reported performance on \dataset is significantly worse than other benchmarks for all models across all metrics (Appendix ~\ref{app:sota_model_perf}). 

\begin{figure}[]
  \centering
  \includegraphics[width=\textwidth]{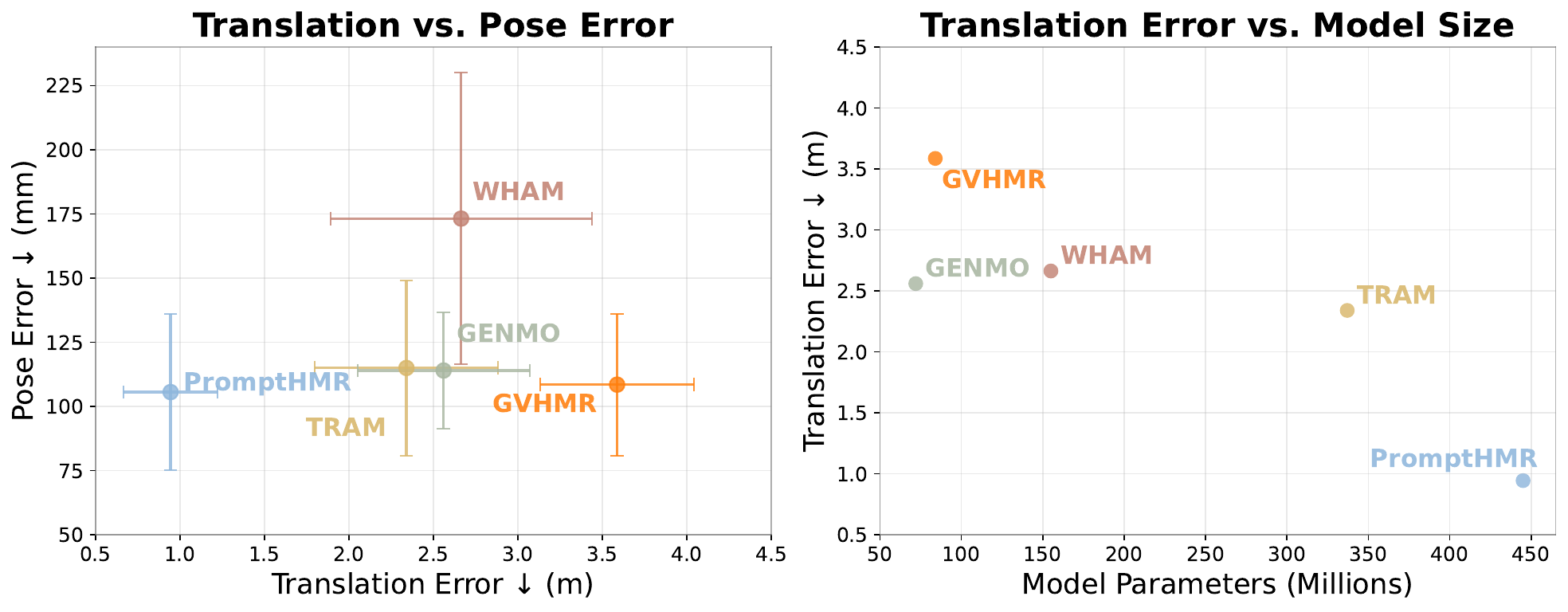}
  \caption{
  \textbf{Multi-View Consistency Analysis.} {\bf (Left)}:  Median pose error versus translation error ($E_{\text{trans}}$) ($E_{\text{pose}}$) across SOTA models on \dataset. Metrics represent cross-view disagreement (\S\ref{sec:metrics}); error bars denote 25th/75th percentiles. {\bf (Right)}: Translation error versus model size (num. parameters). PromptHMR has the lowest translation inconsistency but also the heaviest, while TRAM, which is second lowest, is also second lightest.}
  \label{fig:model_perf}
\end{figure}

\begin{figure}[t]
  \centering
  \includegraphics[width=\textwidth]{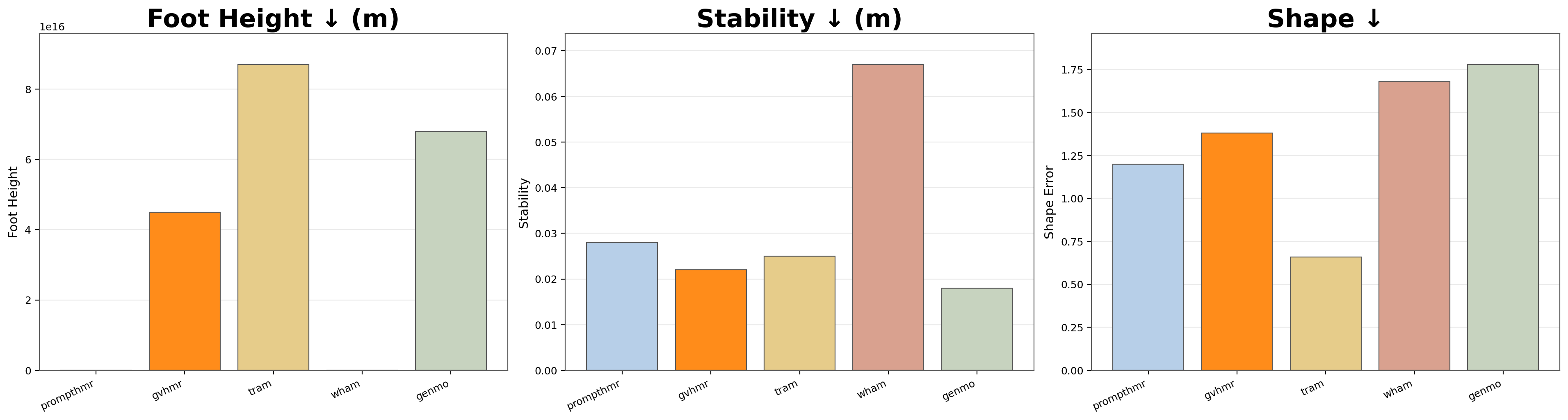}
  \caption{\textbf{Consistency Across Motion Metrics.} Multi-view agreement for foot height ($E_{\text{height}}$), stability ($E_{\text{stab}}$), and shape estimates (\S\ref{sec:metrics}). No single model dominates across all dimensions; e.g., while WHAM excels in foot height consistency, it shows high inconsistency in stability and shape metrics. Results highlight the trade-offs between static pose accuracy and temporal/physical consistency (\S\ref{sec:overall_performance}).}
  \label{fig:metrics_bar_plots}
\end{figure}

In Figure \ref{fig:model_perf} we compare translation and pose error more closely, plotting median values of each metric, with error bars indicating the 25th and 75th percentiles.  We find that all models struggle with making consistent translation estimates, with average error ranging from 0.9m - 3.6m, with 75\% of translation errors within a 1m window. As the poses contained in \dataset span greater distances, pixel-level errors in pose estimates can result in more severe mistakes in translation estimates, an effect that is not obvious from current benchmarks. Qualitatively, we find that this results in a "pose drifting" effect, or oscillations in translation estimates along each camera's depth axis. Models are much more consistent when it comes to pose estimates, with about 11cm error between multi-view poses across all models. This suggests that these models are more ready for downstream applications involving pose estimates alone, rather than those dependent on accurate 3D identification of people in the scene. In Figure \ref{fig:metrics_bar_plots} we compare the consistency of different models along foot skating, stability, and shape estimates. In addition to there being no best model across the board, the relative ordering of model performance for each metric changes as well.

\begin{figure}[]
  \centering
  \includegraphics[width=\textwidth]{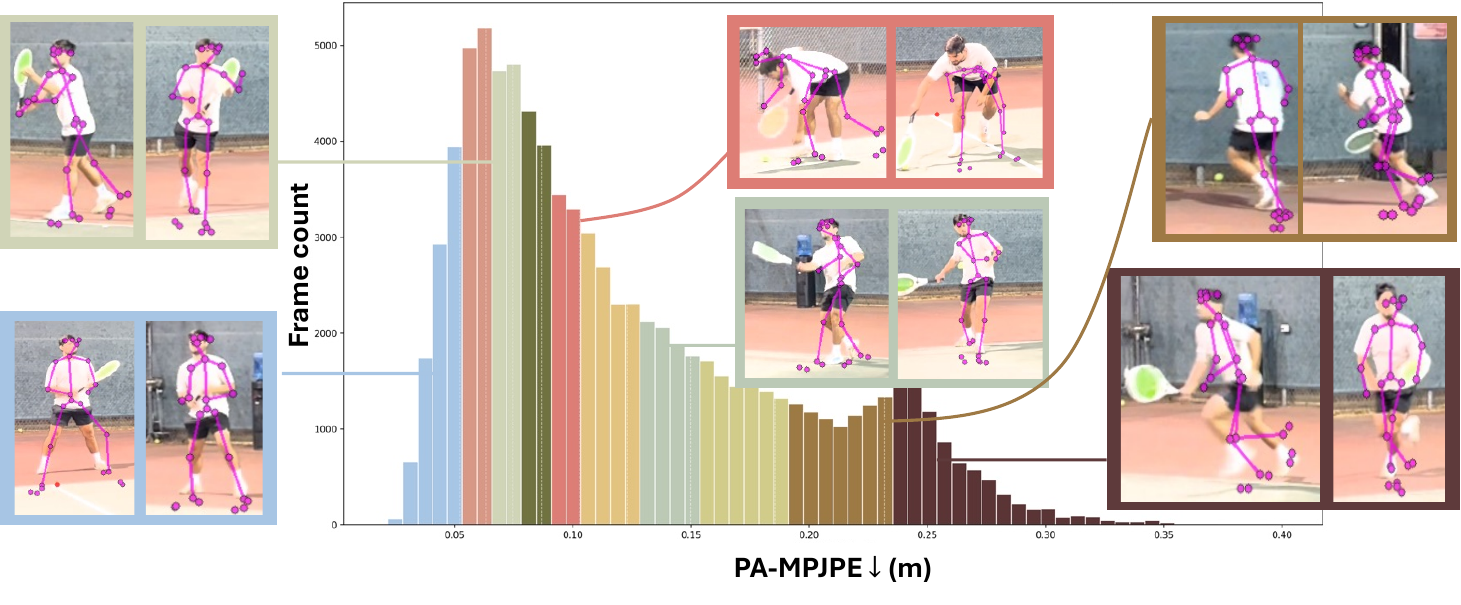}
  \caption{\textbf{Cross-view pose projections.} 
  Histogram of multi-view PA-MPJPE inconsistency for PromptHMR (the best-performing model) on a single video, colored by 10\% intervals. We project pose estimates from one camera onto the other view to highlight inconsistencies. Low-disagreement poses are typically stationary and equally visible by both cameras, while high disagreement occurs on distant or dynamic poses with some occlusion.
  }
  \label{fig:qualitative_fig}
\end{figure}

We provide a qualitative example of a two-view video in Figure \ref{fig:qualitative_fig}. We find that multi-view error tends to be normally distributed. High-disagreement frames often correspond to poses in motion, or in which the depth scale from one view is unclear. Low-disagreement frames typically correspond to stationary, "neutral" poses that are clearly resolved from both views. Across all examples, poses disagree most in the positions of feet and hands, especially when there are partial occlusions.

\subsection{Pose bias}

\begin{figure}[]
  \centering
  \includegraphics[width=\textwidth]{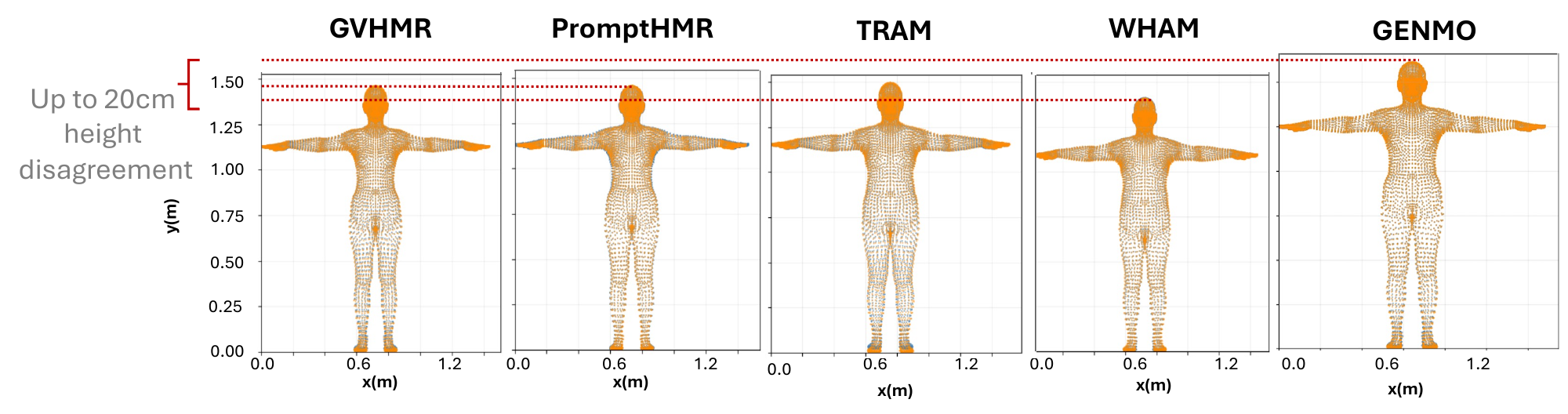}
  \caption{\textbf{Shape consistency.} 
  Models exhibit significant inconsistency in SMPL-X shape parameters ($\beta$) across different views. Qualitatively, PromptHMR (\S\ref{sec:overall_performance}) achieves the highest multi-view consistency, likely due to its conditioning on 2D bounding boxes and keypoints.
  }
  \label{fig:shape_consistency}
\end{figure}

A metric that is overlooked and yet affects all downstream motion analysis is shape consistency, or a person's height, weight, and musculature represented by the SMPL-X shape parameter. We show a qualitative example of model shape consistency in Figure \ref{fig:shape_consistency}. Notably, all models produce inconsistent predictions across different views, suggesting models are sensitive to camera angle. PromptHMR produces the most consistent multi-view shape reconstruction, likely because it takes in additional bounding box and joint information. There are even greater discrepancies across different models, suggesting that each has its own learnt pose bias. Furthermore, in order to produce consistent trajectories, models often predict shape parameters once per video, based on the first frame. Our findings suggest the need for a more nuanced video-level approach to shape prediction.

\section{Discussion and Conclusions}
\label{sec:conclusions}


We introduce \dataset, a large-scale, multi-view video dataset of real-world tennis practices and match play, alongside a label-free evaluation framework that uses multi-view consistency to lower-bound monocular reconstruction error. Across eleven million frames of unscripted athletic motion, the dataset exposes monocular 3D pose estimators to depth ranges, pose variability, and action repetition that existing in-the-wild benchmarks do not cover, without requiring MOCAP, body-worn sensors, or manual annotation --- only consumer phones, inexpensive tripods, and the geometry of a tennis court.

Our benchmark findings answer two questions. First, is any one model best? No. PromptHMR leads on conventional translation and joint-position metrics but shows the largest per-video variance; WHAM collapses foot skating and pose drift to near zero at the cost of much worse translation accuracy; GENMO is the most internally consistent across views; PromptHMR produces the most stable body shape across views. No model dominates across the axes that matter.
Second, is the best available model good enough for downstream applications? Also no --- but the answer depends on the application. Joint-angle pose recovery is now accurate enough that tasks depending primarily on relative body configuration and temporal kinematics --- activity recognition, coarse technique analysis, and gait-based identification --- can proceed with care. But three quantities remain unreliable across every model we tested: absolute distance and depth, ground-contact detection, and body shape (limb lengths, height, proportions). These are exactly the quantities that clinical biomechanical analysis, force and balance estimation, fine-grained sports analytics, pedestrian intent prediction, and forensic stride-length measurement most directly depend on. A coach reading stroke technique from a phone video can trust current models more than a clinician reading weight transfer, and far more than a biomechanist measuring ground reaction forces from video alone.

Beyond these findings, our multi-view consistency framing is portable. Any benchmark that can be captured from multiple views inherits a label-free error signal, which we hope lowers the cost of building the next generation of in-the-wild benchmarks in other sports, clinical motion analysis, and everyday activity. We release dataset, capture recipe, and evaluation code to make this path concrete.



\paragraph{Limitations \& Broader Impacts.}
CalTennis is currently limited to a single sport, climate, and surface type, captured by one research group. Our "easy and inexpensive to collect" claim has not yet been verified through replication by other teams. Crucially, multi-view disagreement provides a \emph{lower} bound on error, making our dataset a complement to, rather than a replacement for, absolute MOCAP validation. While \dataset aims to advance accessible motion analysis, large-scale human datasets carry inherent privacy and surveillance risks; we mitigate these via IRB approval, informed consent, and mandatory face blurring. Finally, our benchmark reveals that current monocular models are unreliable when estimating depth, foot contact, and body shape, making them unsuitable for most clinical and forensic settings.


\newpage
\section{Acknowledgments and Disclosure of Funding}

We would like to thank Damiano Marsili, Aadarsh Sahoo, and Dylan Zhou for valuable feedback and discussion, as well as the Caltech tennis coaches, Adam Clark and Rachel Viollet. This work is supported by the Technology Innovation Institute (TII) and the National Science Foundation Graduate Research Fellowship Program under Grant No. 2139433. Data of the Caltech Tennis Team was obtained through explicit IRB approved consent.

 

\clearpage
{\small
\bibliographystyle{plain}
\bibliography{egbib}
}



\appendix
\newpage
\section{Technical appendices and supplementary material}

\subsection{Maximum-likelihood consensus pose}
To establish a single robust 3D joint estimate per timestep from $N$ overlapping views, we form an MLE-weighted consensus pose. Let $J_{c}^{(i)} \in \mathbb{R}^3$ represent a joint position estimated by the monocular model from camera $i$ in its local coordinate frame. We model the per-camera measurement noise as a Gaussian distribution centered at the true camera-frame joint position. To account for the fact that depth is the dominant error mode in monocular reconstruction, the covariance matrix $\Sigma_c^{(i)} \in \mathbb{R}^{3 \times 3}$ is elongated along that camera's depth axis.

Before fusing the estimates, we must lift each camera's prediction into the shared world coordinate system (the tennis court frame). Given the camera-to-world calibration parameters $R_{\text{calib\_c2w}}^{(i)} \in \mathbb{R}^{3 \times 3}$ and $T_{\text{calib\_c2w}}^{(i)} \in \mathbb{R}^3$, the lifted world-frame joint observation is:
\begin{equation}
    J_{w}^{(i)} = R_{\text{calib\_c2w}}^{(i)} J_{c}^{(i)} + T_{\text{calib\_c2w}}^{(i)}
\end{equation}

To ensure correct coordinate transformation, the noise covariance matrix must also be transformed into the world frame. The world-frame covariance $\Sigma_w^{(i)}$ is computed as:
\begin{equation}
    \Sigma_w^{(i)} = R_{\text{calib\_c2w}}^{(i)} \Sigma_c^{(i)} \left(R_{\text{calib\_c2w}}^{(i)}\right)^T
\end{equation}

Let the true consensus world-frame joint position be $P$. Assuming the measurement errors across the $N$ cameras are independent, the joint likelihood of observing the set of lifted estimates $\mathcal{D} = \{J_{w}^{(1)}, \dots, J_{w}^{(N)}\}$ is the product of the individual Gaussian likelihoods:
\begin{equation}
    L(P | \mathcal{D}) = \prod_{i=1}^{N} \frac{1}{\sqrt{(2\pi)^3 |\Sigma_w^{(i)}|}} \exp\left(-\frac{1}{2} (J_{w}^{(i)} - P)^T (\Sigma_w^{(i)})^{-1} (J_{w}^{(i)} - P)\right)
\end{equation}

The Maximum Likelihood Estimate (MLE) for the true pose, $P_{\text{MLE}}$, is found by minimizing the negative log-likelihood:
\begin{equation}
    -\ln L(P | \mathcal{D}) = \frac{1}{2} \sum_{i=1}^{N} (J_{w}^{(i)} - P)^T (\Sigma_w^{(i)})^{-1} (J_{w}^{(i)} - P) + C
\end{equation}

Taking the derivative with respect to $P$ and setting it to zero yields:
\begin{equation}
    \frac{\partial (-\ln L)}{\partial P} = - \sum_{i=1}^{N} (\Sigma_w^{(i)})^{-1} (J_{w}^{(i)} - P) = 0
\end{equation}

Solving for $P$ yields the closed-form solution, which is a precision-weighted average of the lifted per-view estimates:
\begin{align}
    \left( \sum_{i=1}^{N} (\Sigma_w^{(i)})^{-1} \right) P &= \sum_{i=1}^{N} (\Sigma_w^{(i)})^{-1} J_{w}^{(i)} \\
    P_{\text{MLE}} &= \left( \sum_{i=1}^{N} (\Sigma_w^{(i)})^{-1} \right)^{-1} \sum_{i=1}^{N} (\Sigma_w^{(i)})^{-1} J_{w}^{(i)}
\end{align}

This formulation ensures that the highly uncertain depth axes from individual monocular cameras are appropriately down-weighted during spatial fusion.

\subsection{Camera calibration}
\label{app:camera_calib}

\begin{table}[h]
\centering
\caption{Notation and Definitions for Problem Setup}
\label{tab:notation}
\begin{tabular}{ll}
\toprule
\textbf{Variable} & \textbf{Definition} \\
\midrule
$N$ & Number of cameras capturing the scene \\
$c^i$ & The $i$-th camera \\
$V^i$ & Video collected by camera $c^i$, containing RGB frames $I_t^i$ \\
$t^i_k$ & Length (total frames or duration) of video $V^i$ \\
$K^i$ & Intrinsic matrix of camera $c^i$ ($\mathbb{R}^{3 \times 4}$) \\
$R^i, T^i$ & Extrinsic rotation ($SO(3)$) and translation ($\mathbb{R}^3$) of camera $c^i$ \\
$p$ & Number of people in the scene \\
$H^i$ & SMPL-X pose estimates for $p$ people from video $V^i$ \\
$\tau^i_t$ & Translation of all $p$ people at time $t$ ($\mathbb{R}^{p \times 3}$) \\
$\phi^i_t$ & Global orientation of all $p$ people at time $t$ ($\mathbb{R}^{p \times 3}$) \\
$\theta^i_t$ & Body pose (joint angles) of all $p$ people at time $t$ ($\mathbb{R}^{p \times 21 \times 3}$) \\
$\beta^i_t$ & Body shape parameters of all $p$ people at time $t$ ($\mathbb{R}^{p \times 10}$) \\
$M$ & Pre-trained monocular 3D human pose estimation model \\
\bottomrule
\end{tabular}
\end{table}


In order to run multi-view label-free evaluations of state-of-the-art human pose estimation models, we lift pose predictions from each camera into a shared global space-time reference frame, by calibrating cameras and synchronizing the videos. We describe our method in this section, and in further detail in ~\ref{app:camera_calib}.


\paragraph{Problem definition.}
Given a set of $N$ cameras facing the same scene, each camera $c^i$ collects a video $V^i = \left\{ I_t^i \right\}_{t=0}^{t^i_k} \in \mathbb{R}^{H \times W \times 3}$ of $t^i_k$ RGB frames. We represent the geometry of each camera $c^i$ as an intrinsic matrix  $K^i \in \mathbb{R}^{3 \times 4} $, extrinsic rotation $R^i \in SO(3) $, and extrinsic translation $T^i \in \mathbb{R}^3$. 
We define the poses of $p$ people in each video $V^i$ at time $t$ via SMPL-X\cite{pavlakos2019smplx} parameters $ H^i = \{ (\tau^i_t, \phi^i_t, \beta^i_t, \theta_t^i) \}_{t=t_0^i}^{t^i_k} $, with representing $\tau^i_t,  \in \mathbb{R}^{p \times 3}  $ translation, $\theta_i^t \in \mathbb{R}^{p\times21\times3}$ body pose, $\phi_i^t \in \mathbb{R}^{p \times 3} $ orientation, and $\beta_i^t \in \mathbb{R}^{p \times 10}$  body shape of all $p$ people in the scene. As shown in Fig.\ref{fig:time-calib}, given 3D human pose estimates $ H^i = M(V_i)$ generated by a pre-trained model $M$ run on each video $V_i$, we seek to lift all 3D pose estimates $\{ H^0, ..., H^N \}$ into a shared spatio-temporal world coordinate system in order to evaluate the multi-view consistency of model predictions. An additional challenge comes from varying wall clock times. For example, two cameras $c^1$ and $c^2$ will have distinct timestamps $ \{ t_0^1, ... t_k^1 \} $ and $ \{ t_0^2, ... t_k^2 \} $, even if they have the same frame rate. Furthermore, camera extrinsics $(R^i, T^i)$ are not known a priori, so we cannot directly compare model outputs even if they are in a metric-scale coordinate system. Thus, we perform temporal calibration, calibrate cameras, and lift human pose estimates into shared global world coordinates. We discuss these three steps in further detail in the following sections.


\paragraph{Video recording setup and camera calibration.}
\label{sec:cam_calib}
We aim to capture human tennis motions from views that cover all sides of the player while encompassing the full range of the tennis court. To achieve this, we position $N$ (up to 6) cameras around the court (Figure \ref{fig:dataset-overview}). Since a player's motions are typically constrained to one half of the court and monocular model predictions degrade significantly over distances exceeding half the court length, we orient cameras toward each half-court independently. We calibrate each camera $c^i$ by leveraging the standardized dimensions of the tennis court. Specifically, for a given frame $I_t^i$ from video $V_i$ captured by camera $c^i$, we identify a set of $n$ court line intersections $\{ P^i_0, ...,  P^i_n \} \in \mathbb{R}^3 $ and their corresponding detected pixel coordinates $ \{ p^i_0, ... p^i_n \} \in \mathbb{R}^2 $. We obtain the camera intrinsics $K^i$ from the iPhone metadata and define the camera projection function $\pi(X; K^i): \mathbb{R}^3 \rightarrow \mathbb{R}^2$ which projects $X \in \mathbb{R}^3$, a point in world coordinates down to $x \in \mathbb{R}^2$, a point in pixel coordinates. The extrinsic rotation $R^i$ and translation $T^i$ are recovered by minimizing the reprojection error:
\begin{equation}
    \min_{R^i, T^i} \sum_{k=1}^{n} \left\| \pi(R^i {P}_k + T^i; K^i) - {p}_k \right\|^2
\end{equation}

\begin{figure}[]
  \centering
  \includegraphics[height=4cm]{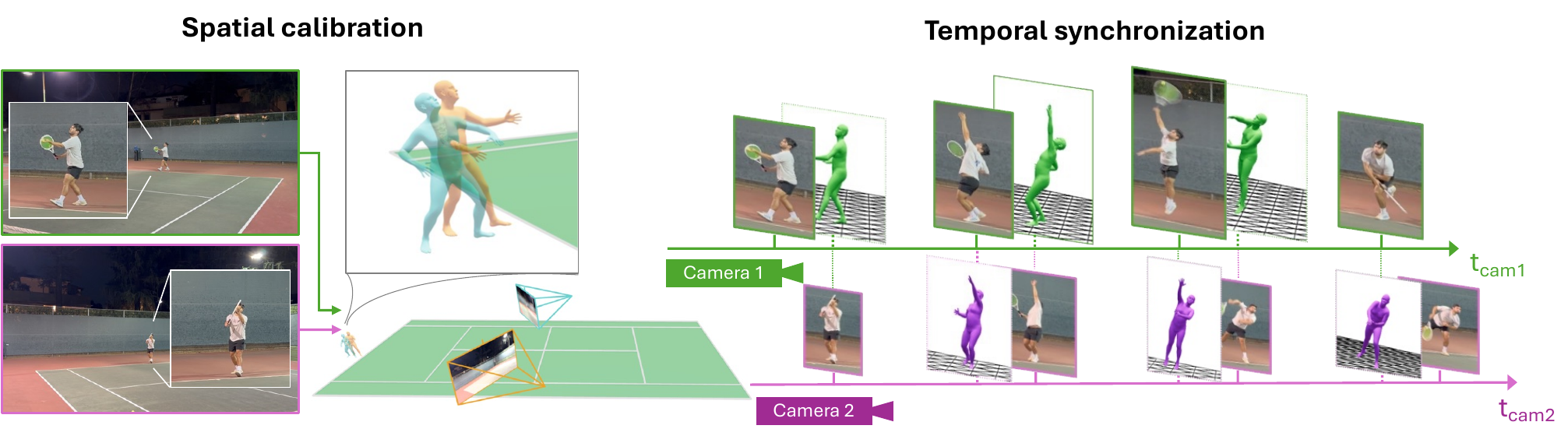}
  \caption{\textbf{Spatiotemporal calibration.}
  Left: We lift model estimates into a shared court coordinate system (\S\ref{sec:spat_fusion}). Discrepancy in depth estimates results in differing 3D translation estimates $\tilde{\tau}_t^i$. \\
  Right: Videos lack identical timestamps, so we align sequences using a global offset $\Delta t$ and linearly interpolate poses for missing timestamps (\S\ref{sec:temp_calib}) to ensure a precise millisecond-level comparison.
  }
  \label{fig:time-calib2}
\end{figure}

We implement this using the PnP algorithm.

\paragraph{Spatial fusion}
\label{sec:spat_fusion}

Now, equipped with per-camera pose estimates $ H^i = \{ (\tau^i_t, \phi^i_t, \beta^i_t, \theta_t^i) \} $ for camera $c^i$ with intrinsics $K^i$ and extrinsics $(R^i, T^i)$, we lift these poses into a shared world coordinate system. We define the transformation from model-space to world-space as $T_{W_i^{\text{model}} \to W}$, and we can express it as:
\begin{equation}
  T_{W_i^{\text{model}} \to W}
  =
  \begin{bmatrix} {R^i}^\top & -{R^i}^\top{T^i} \\ \mathbf{0} & 1 \end{bmatrix}
  \label{eq:world_align}
\end{equation}
Eq.~\eqref{eq:world_align} now lets us lift translation estimates $\tau_t^i$ for each person $p$ from camera coordinates into the global world coordinate system via:
\begin{equation}
  \tilde{\tau}_t^i = T_{W_i^{\text{model}} \to W} \, [\tau_t^i; 1].
\end{equation}

We additionally formulate a consensus pose by obtaining an Maximum Likelihood Estimate between all poses captures by cameras $C_i$. Depth estimation is the biggest source of error in monocular 3D human pose reconstruction (Section \ref{sec:overall_performance}), so we model the measurement noise for camera $c$ as a multivariate Gaussian $\mathcal{N}(\mathbf{0}, \Sigma_c)$, where the covariance $\Sigma_c$ is elongated along the camera's depth axis. The fused global pose $\hat{p}^t$ that maximizes the log-likelihood across all $C$ cameras is:
\begin{equation}
    \hat{\mathbf p}_i =
    \left( \sum_{j=1}^{C} \tilde{\Sigma}_j^{-1} \right)^{-1}
    \left( \sum_{j=1}^{C} \tilde{\Sigma}_j^{-1} \tilde{\mathbf p}_i^j \right).
\end{equation}
In which we have transformed the noise covariance estimate $\Sigma_j$ for camera $C_j$ using $T_{W_i^{\text{model}} \to W}$. We provide a full derivation of the MLE estimate in the Appendix.



\paragraph{Temporal calibration.}
\label{sec:temp_calib}
Athletes move at high speeds; modeling athletic motion requires fine-grained time estimates. After running a monocular pose reconstruction model on each view, we obtain 3D human pose estimates from a pre-trained model $M$ on each video $V^i$, where $ M(\{ V^i \}) = \{  \tau^i_t, \theta^i_t, \phi^i_t, \beta^i_t  \} $ for all frames taken by camera $C_i$. 
However, we do not have identical time estimates across videos, so we do not have temporally synchronized poses. For example, in the case of two cameras as shown in Fig.\ref{fig:time-calib}, we do not have pose estimates for the video taken by camera 2 (purple) at the timestamps of camera 1 (green) and vice versa. 
We could compare poses between closest nearby timestamps, but this entangles temporal shift from model-based errors, and in sports motions exhibit changes on the millisecond-level. Thus, we must estimate a pose at time $t_i$ for the pose reconstructed from camera $C_j$ by linearly interpolating between the pose at nearest two time estimates around timestamp $t_i$.
Additionally, iPhones record wall clock time up to the nearest second rather than millisecond in their video recording, so there may be up to a 1000-millisecond discrepancy between global timestamps of different videos. We address this issue by estimating an optimal offset factor $\Delta t_c^t$ from a grid search over a range between -1000 ms and 1000 ms that minimizes the pose disagreement between reconstructed pose estimates. 
We provide an example calibration in Figure \ref{fig:time-calib}.



\subsection{Are difficult frames difficult for everyone?}

\begin{figure}[]
  \centering
  \includegraphics[width=\textwidth]{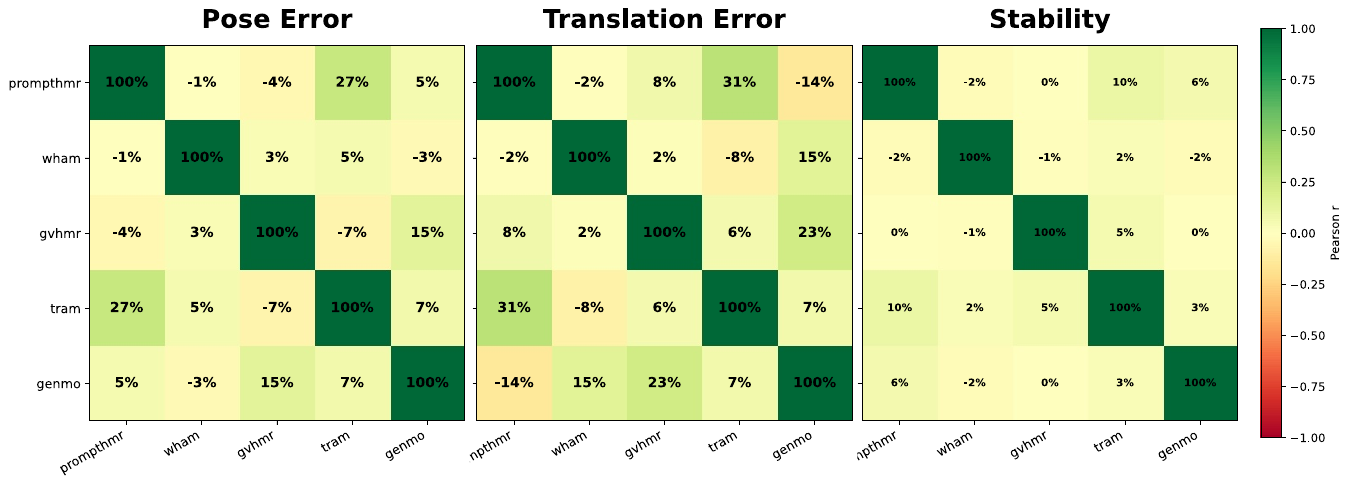}
  \caption{\textbf{Model Error Correlations}. We calculate frame-level Pearson correlation between error measurements of different models for pose error, translation error, and stability. Each square shows the correlation between the error (in time) signal of two models. We find little correlation in model error across different models.
  }
  \label{fig:model_err_corss}
\end{figure}

Do models find the same set of motions difficult? We investigate this in several ways. We report the average per-frame Pearson correlation between different models in Figure \ref{fig:model_err_corss}. We find that there is little correlation across model pose errors than translation errors, suggesting that models fail for different reasons. One exception to this is TRAM, which has a Pearson correlation of .27 PromptHMR for Pose L2 error and .31 for translation error, whereas it only has a .1 correlation when it comes to stability. This is consistent with the deep uncertainty and variation in depth estimates that we find across all models.

\begin{figure}[]
  \centering
  \includegraphics[width=\textwidth]{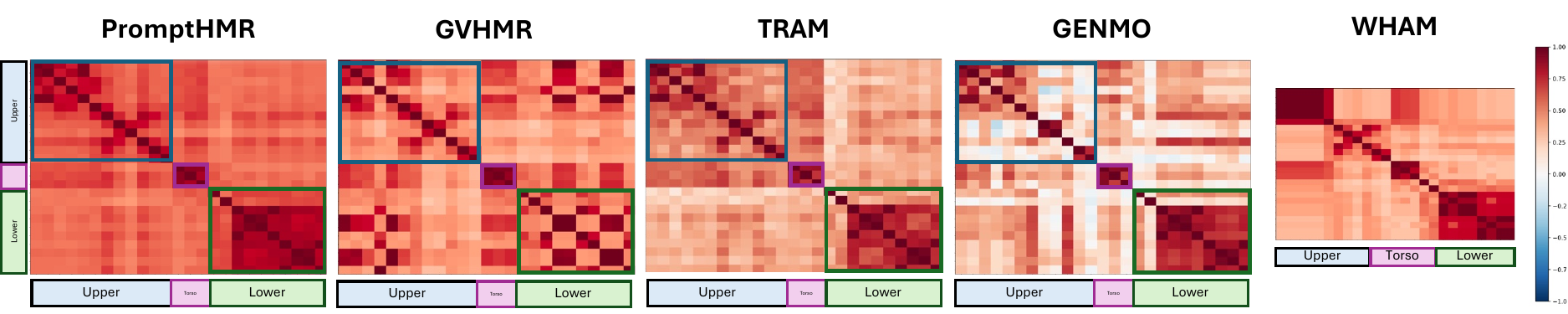}
  \caption{\textbf{Joint Error Correlations}. We notice that upper body (blue box) and lower body (green box) errors do not correlate with each other; erroneous upper-body estimates can correspond to consistent lower-body estimates and vice-versa. WHAM produces SMPL coordinates, which contain fewer upper and lower body joints. Interestingly, the torso joints (left/right hip and pelvis) correlate more with lower body errors for PromptHMR, and correlate much more strongly with upper body errors for GVHMR.
  }
  \label{fig:joint_corr}
\end{figure}

We find an additional commonality in model performance when we inspect joint error correlations, shown in Fig \ref{fig:joint_corr}. We find that error is correlated for upper body joints with themselves, pelvic joints with themselves, and lower body joints with themselves, but there is little correlation in errors between upper and lower body joints. This suggests that unreliable estimates in upper body position tell us little about how much to trust lower body estimates, however if we know that estimates in the wrist are inconsistent then this suggests similar errors in head and shoulder estimates. This finding also has repercussions for how we collect sensor-aided data: using just foot contact points alone or just IMU sensors on the upper body together with a joint pose optimization pipeline is likely to be a poor estimate of the true ground-truth poses.



\subsection{SOTA model performance}
\label{app:sota_model_perf}

In Table \ref{app:sota_model_perf_on_current_benchmarks} we report the performance of state-of-the-art human pose estimation models on the 3DPW~\cite{black20183dpw}, RICH~\cite{black2022rich}, and EMDB~\cite{kaufmann2023emdb} evaluation datasets. PromptHMR~\cite{wang2025prompthmr} performs best on RICH and EMDB on standard pose and translation metrics, but worst on acceleration metrics, consistent with our findings in Section \ref{sec:overall_performance} that it performs most consistently on pose and translation estimates but much worse on foot sliding and stability, dynamic metrics that directly affect acceleration. GENMO performs best on 3DPW, but none of the other benchmarks. 

\begin{table}[]
    \caption{\textbf{SOTA Model Performance on Current Benchmarks.} We report the performance reported by each paper, on the standard human motion evaluation metrics.}
    \label{app:sota_model_perf_on_current_benchmarks}
    \centering
    \small
    \resizebox{\textwidth}{!}{
    \setlength{\tabcolsep}{4pt} 
    \begin{tabular}{ll|cccc|cccc|cccc}
        \toprule
        & & \multicolumn{4}{c}{3DPW~\cite{black20183dpw}} & \multicolumn{4}{c}{RICH~\cite{black2022rich}} & \multicolumn{4}{c}{EMDB~\cite{kaufmann2023emdb}} \\
        
        \multicolumn{2}{l|}{Models} & PA-MPJPE & MPJPE & PVE & Accel & PA-MPJPE & MPJPE & PVE & Accel & PA-MPJPE & MPJPE & PVE & Accel \\
        \midrule
        
        \parbox[t]{2mm}{\multirow{6}{*}{\rotatebox[origin=c]{90}{per-frame}}} 
        
        & WHAM~\cite{shin2024wham}
            & 37.2 & 59.4 & 71.0 & 6.9
            & 44.7 & 82.6 & 93.2 & 5.6
            & 48.8 & 80.7 & 93.7 & 5.9 \\
        & GVHMR~\cite{xiaowei2024gvhmr}   
            & 36.2 & 55.6 & 67.2  & 5.0
            & 39.5 & 66.0 & 74.4 & \textbf{4.1}  
            & 42.7 & 72.6 & 84.2 & \textbf{3.6}   \\
        & TRAM~\cite{daniilidis2024tram}    
            & 35.6 & 59.3 & 69.6 & \textbf{4.9}
            & - & - & - & - 
            & 45.7 & 74.4 & 86.6 & 4.9 \\
        & PromptHMR~\cite{wang2025prompthmr}  
            & 35.5 & 56.9 & 67.3  & -- 
            & \textbf{37.0} & \textbf{57.4} & \textbf{65.8} & --   
            & \textbf{40.1} & \textbf{68.1} & \textbf{79.2} & --   \\
        & GENMO~\cite{yuan2025genmo}   
            & \textbf{34.6} & \textbf{53.9} & \textbf{65.8}  & 5.2
            & 39.1 & 66.8 & 75.4 & \textbf{4.1}
            & 42.5 & 73.0 & 84.8 & 3.8 \\
        \bottomrule
    \end{tabular}%
    }
\end{table}

\subsection{Additional dataset complexity analyses}
\label{app:further_complexity_anlyses}
\begin{figure}[]
  \centering
  \includegraphics[width=\textwidth]{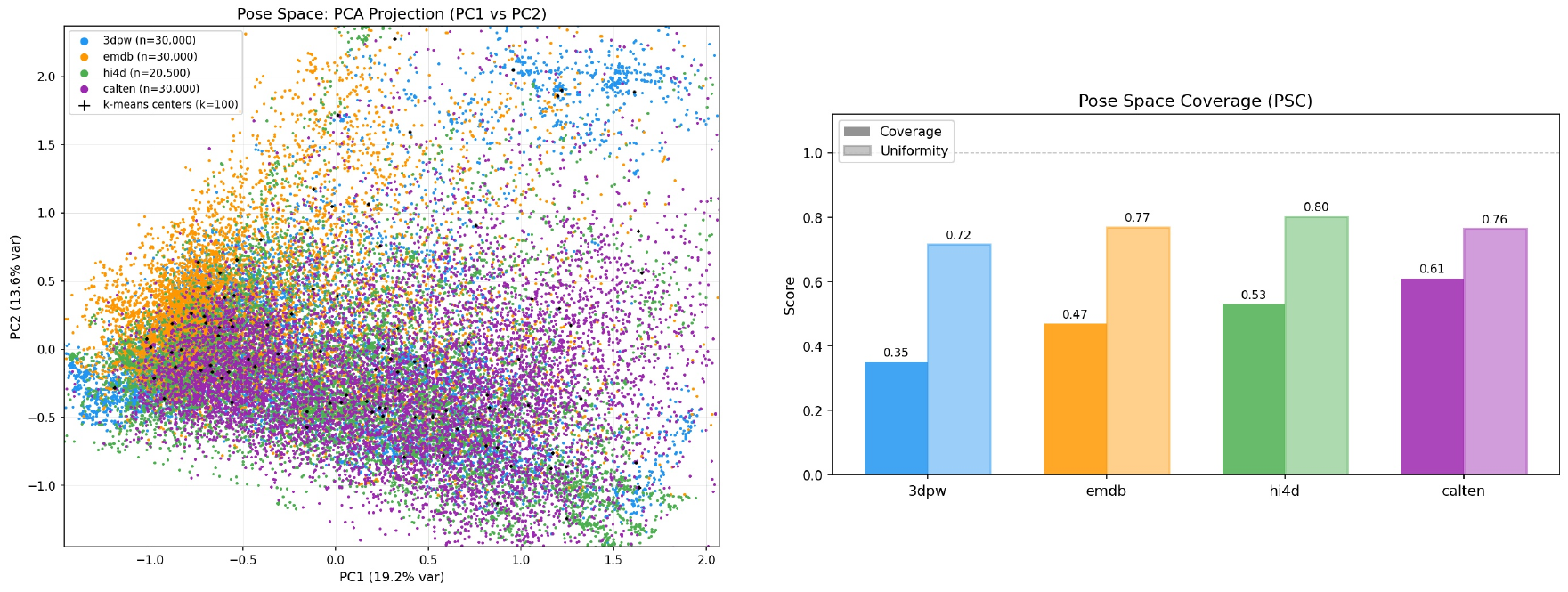}
  \caption{\textbf{Pose Space Uniformity and Coverage of Real-World Datasets}. 
  }
  \label{fig:pose_complexity_2}
\end{figure}

In Figure \ref{fig:pose_complexity_2}, we report additional pose space complexity metrics of current real-world evaluation datasets. We plot the first two dimensions of the PCA decomposition of joint pose space (flattened joint vectors of 3D joint positions) with $k=500$ clusters. \dataset points, shown in purple, are the most evenly spread out throughout the space. Visually, this shows us that there is much more coverage of pose space by \dataset compared to others. On the right we provide a comparison of pose space coverage and uniformity metrics. Coverage is defined as the number of clusters that points in a dataset visit, divided by the total number of clusters (in this case 500). Joint poses in \dataset visit 10\% more clusters than other benchmarks. Uniformity is defined as the ratio of the shannon entropy of the proportion of the poses in a cluster, divided by the maximum possible entropy. Interestingly, uniformity is roughly equal among all the datasets, with HI4D and \dataset within 4 percentage points of each other.

\begin{figure}[]
  \centering
  \includegraphics[width=\textwidth]{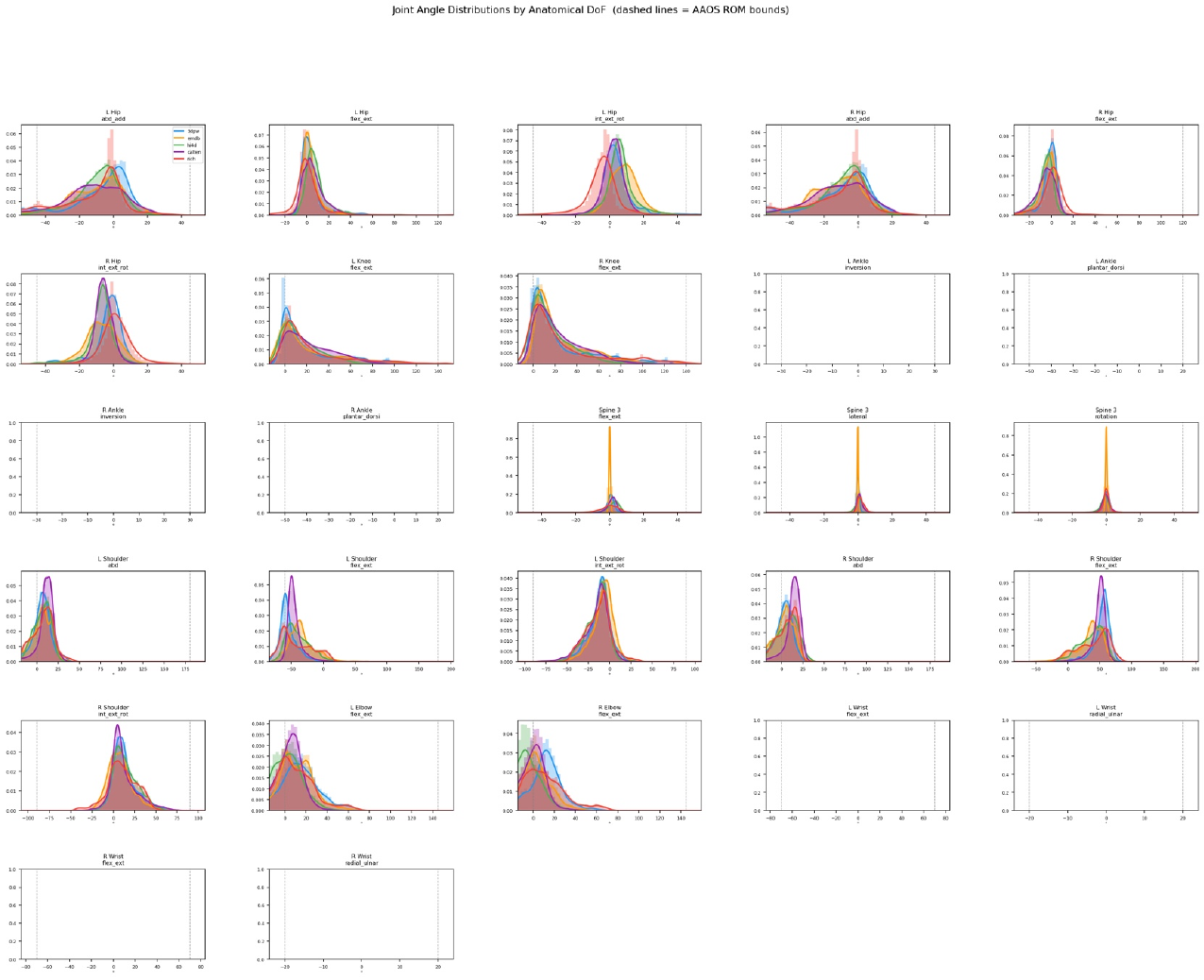}
  \caption{\textbf{Joint Angle Histograms}. We report the per-joint angular ranges in each dataset, ranging from the 10th to 90th percentiles, and normalize this with the documented angular range from medical literature. A flatter distribution indicates a more even spread over angular mobility. 
  }
  \label{fig:joint_error_corrs}
\end{figure}

In Figure \ref{fig:joint_error_corrs} we provide histograms of per-joint angular distributions. We report the per-joint angular ranges in each dataset, ranging from the 10th to 90th percentiles, and normalize this with the documented angular range (from medical literature). This gives us a per-joint score of 0-1, which we average over. A flatter distribution indicates a more even spread over angular mobility. \dataset is the most evenly spread out for hip abductions, external knee rotations, and spinal flexion. EMDB~\cite{kaufmann2023emdb} seems to contain more shoulder ranges and RICH~\cite{black2022rich} more elbow ranges. This could be because people, when prompted to move around a space, swing their arms about arbitrarily. 

\subsection{Additional model performance metrics}

\begin{figure}[]
  \centering
  \includegraphics[width=\textwidth]{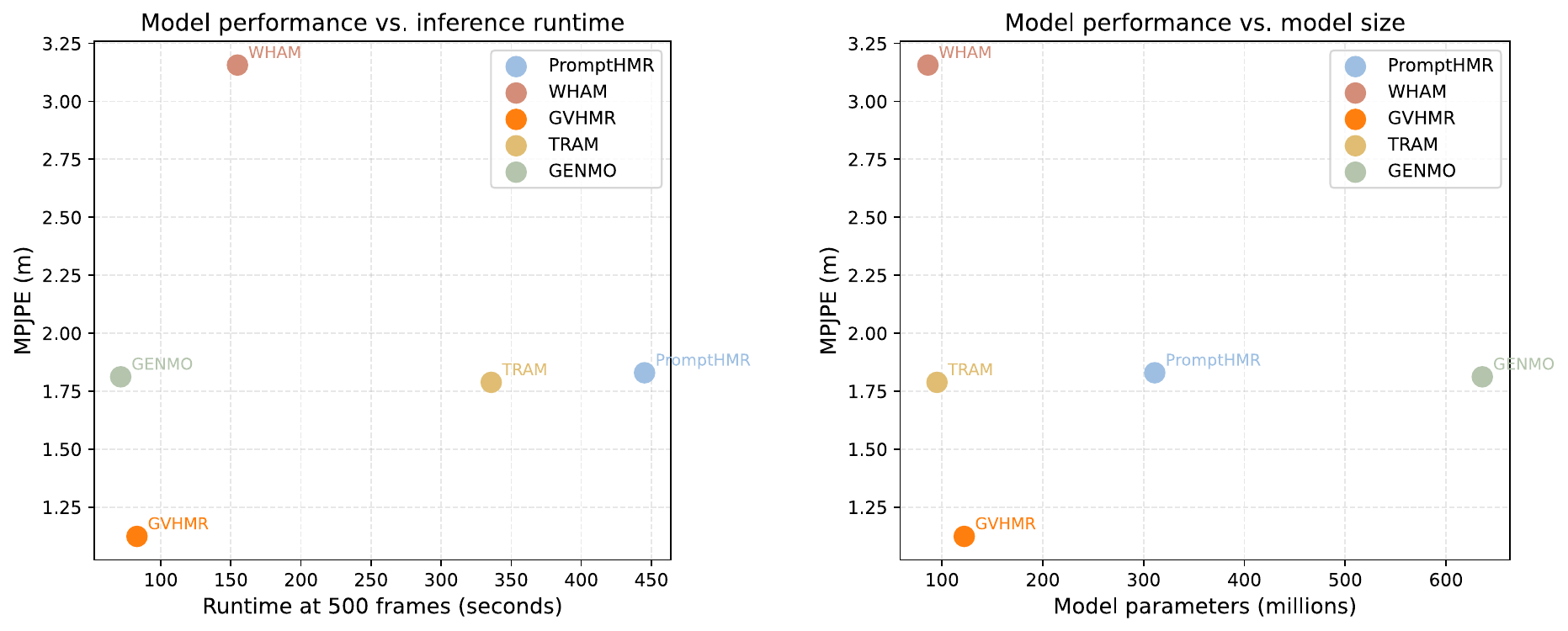}
  \caption{\textbf{Model Performance Trade-Offs}. Left: we plot model runtime versus input frame counts of the same video. PromptHMR seems to scale quadratically, due to having a temporal transformer module, and GENMO scales linearly, thanks to its diffusion-based architecture. Right: we plot mean translation error versus inference runtime.
  }
  \label{fig:model_perf_tradeoffs_2}
\end{figure}

In Figure \ref{fig:model_perf_tradeoffs_2} we report additional model performance comparisons. On the left we plot the model runtime versus MPJPE (mean per-joint position error) consistency. On the right we plot model parameters versus MPJPE. We find that GVHMR~\cite{xiaowei2024gvhmr}, which does not perform best overall \ref{sec:overall_performance}, has the best runtime performance. In other words, it has an optimal tradeoff when it comes to running quickly and accurately. On the right we plot the mean translation error versus the model runtime.

\subsection{Runtime analysis}

\begin{figure}[]
  \centering
  \includegraphics[width=\textwidth]{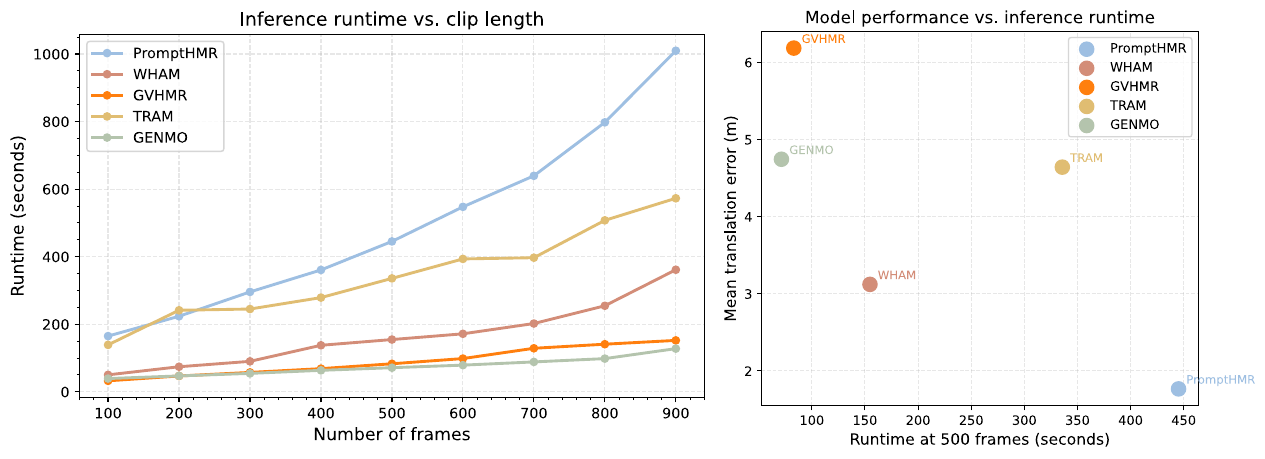}
  \caption{\textbf{Model Runtime Analysis}. 
  }
  \label{fig:runtime_analysis}
\end{figure}

In Figure \ref{fig:runtime_analysis} we investigate model inference time speeds. Unsurprisingly, PromptHMR, the model that we found most consistent in its pose predictions (\ref{sec:overall_performance}), has the slowest runtime, and furthermore scales more steeply with the number of frames in the input video. We expect this is due to the fact that it has a heavy temporal processing step on top of its per-frame processing module. Unsurprisingly, we find that PromptHMR, which is the most performant, is also the slowest. For applications where accuracy matters over speed, PromptHMR is a likelier candidate.

\subsection{Camera setup}

\begin{figure}[]
  \centering
  \includegraphics[width=\textwidth]{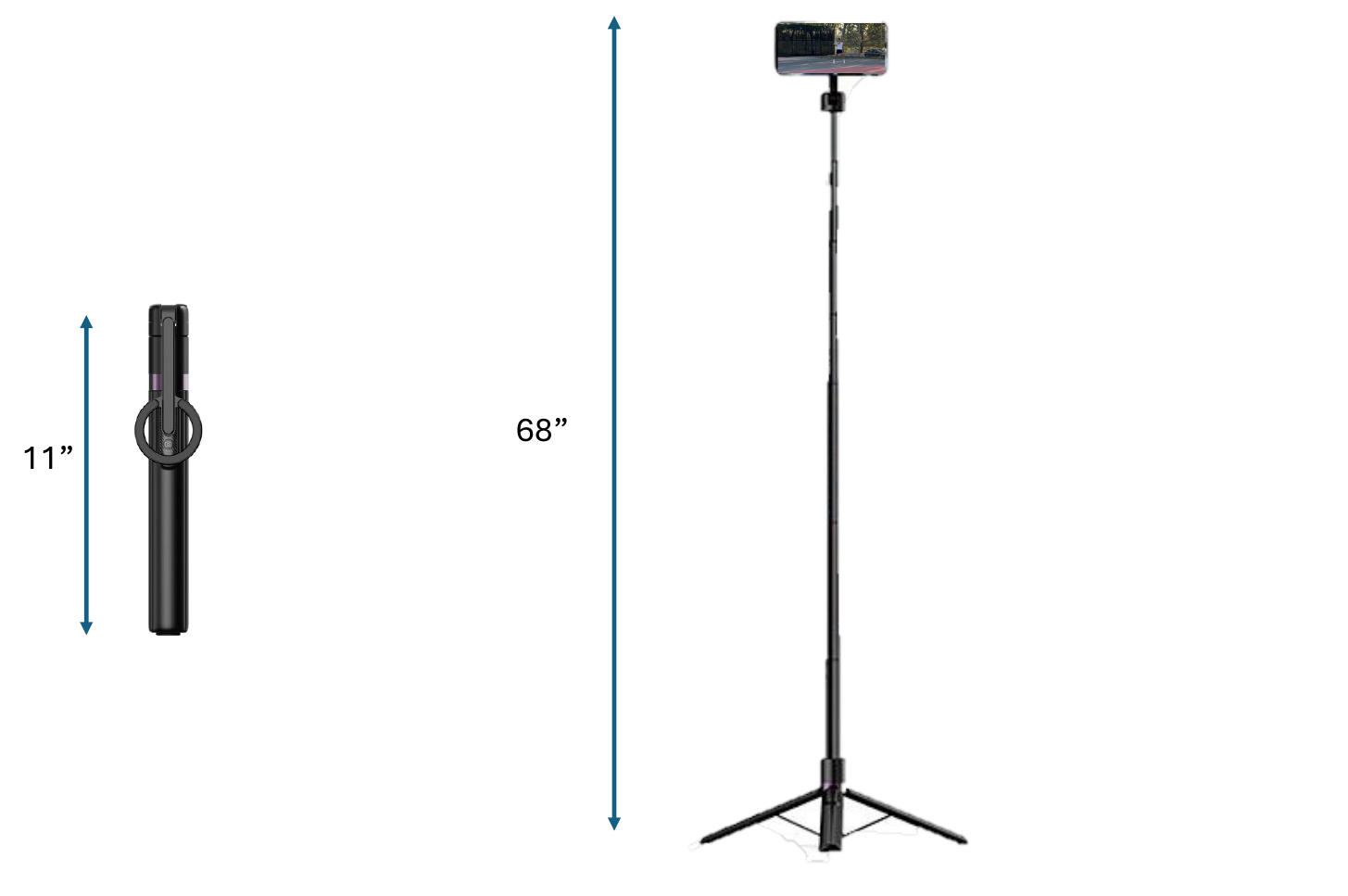}
  \caption{\textbf{Tripod setup}. 
  }
  \label{fig:tripod_setup}
\end{figure}

In Figure \ref{fig:tripod_setup} we provide an example tripod iPhone setup. The tripods we use are cheap (costing about \$40), small (11 inches when folded) and lightweight (less than 1lb). We attach them to iPhones, which are commonly used. This setup could also be extended to other commonly used phones. In our \dataset dataset, we collect up to 6 concurrent recordings of a scene.

\subsection{Model error correlations}
\label{app:model_error_correlations}

To understand potential setting properties that cause monocular pose reconstruction models to fail, we define, compute and analyze the relationship between scene-descriptive variables and multi-view consistency errors for each model, and we discover the correlation between each model's failure patterns.

We define a per-model failure indicator based on multi-view consistency error. For each model $m$, a frame $f$ is labeled as a failure if its error $\varepsilon_f^m$ lies in the top 30\% of that model’s error distribution,
where the threshold $\tau_m$ is chosen independently per model such that 30\% of frames are labeled as failures for each model.

The analysis set $F$ consists of all frames for which every model produces a valid prediction; no additional filtering based on difficulty or error magnitude is applied.

We define a consensus failure count for each frame $k_f = \sum_{m \in M} D_f^m \in \{0,\ldots,5\}$,
which counts how many models classify a frame as a failure. Specifically, $k_f$ is defined for all frames in $F$, including frames where no model fails ($k_f = 0$).


For each frame $f \in F$, we extract ten features that describe scene properties $\mathbf{x}_f$
(Table~\ref{tab:scene_features}) from the video and from the corresponding world-coordinate pose estimates. These features are computed for all frames in $F$, independently of their failure labels, and are used as predictors in the subsequent correlation and classification analyses. We refer to these ten variables collectively as scene features, as they describe camera configuration, motion, and appearance properties of each frame.

We report two complementary diagnostics:
(i)~Spearman rank correlation $r_s$ between each scalar feature and $D^m_f$, which
measures the monotonic association between a single scene property and model failure; and
(ii)~the AUC of a $\ell_1$-regularised logistic regression classifier trained to predict
$D^m_f$ from the full feature vector $\mathbf{x}_f$, which measures how well all scene
features jointly predict failures.
We evaluate four error types as the target: translation error (Translation), joint
position error (Pose), Procrustes-aligned joint error (PA-MPJPE), and mean
per-joint position error (MPJPE).


Table~\ref{tab:model_comparison} reports Spearman correlations and logistic-regression
AUC for each model and error metric. Our key findings include:

\noindent\textbf{Depth-dependence drives joint-position failures.}
The strongest predictor of failure across all models is depth-dependence:
depth and camera distance consistently show the highest correlations with MPJPE
($r_s$ up to $+0.54$ for GENMO and $+0.46$ for GVHMR).
Scale proxy is strongly negatively correlated with MPJPE for GVHMR
($r_s = -0.44$) and GENMO ($r_s = -0.44$), indicating that persons appearing
smaller in the image (i.e., farther from the camera) are harder to reconstruct accurately.
This confirms that depth ambiguity is a central failure mode for monocular methods. This is expected in our setting: tennis courts span ${\sim}23\text{m} \times 11\text{m}$, and cameras placed at baseline or sideline positions produce subject-to-camera distances ranging from ${\sim}5$m to ${\sim}30$m within a single session, making depth variation a great challenge for monocular reconstruction.

\noindent\textbf{Translation and joint errors share scene-feature dependence patterns; Procrustes-aligned error does not.}
Translation L2 and MPJPE rows share similar signs and magnitudes for most models
(e.g., GENMO depth: $+0.33$ vs.\ $+0.42$; WHAM depth: $+0.25$ vs.\ $+0.26$),
suggesting that absolute position errors are dominated by translation.
By contrast, PA-MPJPE rows are substantially weaker across all models
(max $|r_s| \leq 0.21$, AUC $\leq 0.72$), indicating that once global translation
is removed, failure frames are far less distinguishable by scene geometry alone (residual pose errors arise from factors not captured by our scene descriptors).

\noindent\textbf{Models differ in failure predictability.}
The AUC scores reveal meaningful variation: GENMO is the most predictable
(MPJPE AUC $= 0.89$), while GVHMR and PromptHMR are intermediate
($0.84$ and $0.82$), and TRAM and WHAM are least predictable ($0.77$ and $0.76$). High AUC for GENMO is consistent with its global trajectory-reasoning design, which creates a tight coupling between subject distance and reconstruction quality. Conversely, TRAM's lower AUC does not imply fewer absolute errors; rather, its errors arise from identity-level failure modes (re-identification failures, trajectory breaks) that appear stochastically and are not predictable from scene geometry. This is consistent with the low stability performance observed in the main evaluation.

\noindent\textbf{Motion features are weak predictors.}
Motion magnitude and acceleration show consistently low correlations ($|r_s| < 0.17$ for all models and metrics), indicating that fast motion alone
does not explain where models fail under our benchmark conditions, likely
because the players move rapidly throughout the session, leaving little between-frame variation to correlate with. This stands in contrast to standard video benchmark assumptions: in sports contexts, constant high-speed activity compresses the motion distribution, and scene geometry rather than motion intensity distinguishes hard frames.

\begin{figure}[]
\centering
\includegraphics[width=\linewidth]{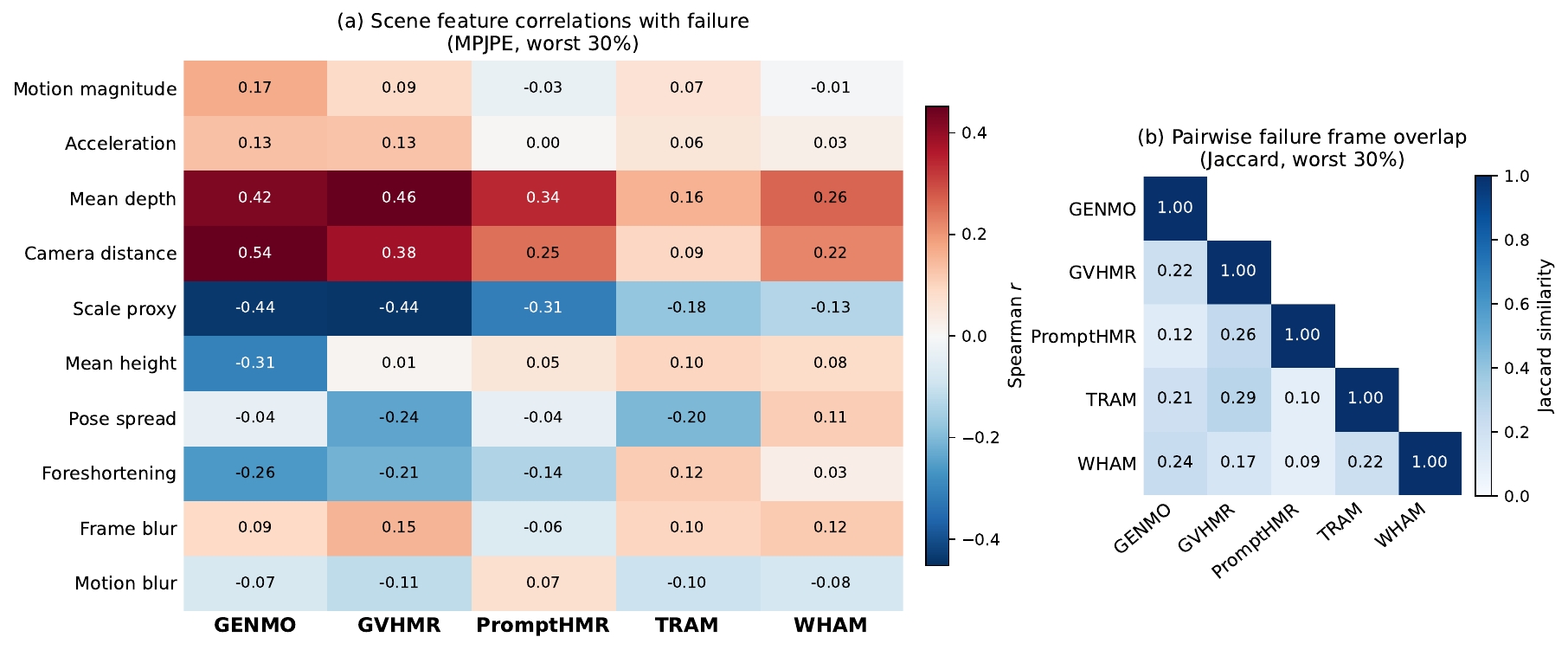}
\caption{Inter-model failure analysis (failure = worst 30\% per model, MPJPE).
  \textbf{(a)} Spearman correlations between scene features and per-model failure labels.
  Geometry features dominate for GVHMR and GENMO; TRAM and WHAM show near-zero signatures.
  \textbf{(b)} Pairwise Jaccard similarity of failure frame sets ($J \leq 0.30$ for all pairs):
  models fail on largely disjoint subsets of frames.}
\label{fig:intermodel}
\end{figure}

\paragraph{Inter-model failure patterns.}
Fig.\ref{fig:intermodel} summarizes inter-model failure structure. We show MPJPE as a representative metric for the heatmap.
Panel~(a) shows the feature-correlation heatmap for MPJPE, which reveals that models carry distinct failure properties. GVHMR and GENMO failures are strongly tied to scene
geometry (depth, camera distance, scale), while TRAM and WHAM show near-zero correlations, indicating their errors are not driven by the scene conditions. These properties are consistent with architectural differences. Specifically, GENMO and GVHMR encode stronger depth-to-error coupling, while TRAM distributes errors more uniformly.
Panel~(b) shows pairwise Jaccard similarity ($J = |A\cap B|/|A\cup B|$) of failure frame sets. All pairs have $J \leq 0.30$, confirming failures are
largely model-specific rather than driven by shared hard scenes.

\begin{figure}[h]
\centering
\includegraphics[width=\linewidth]{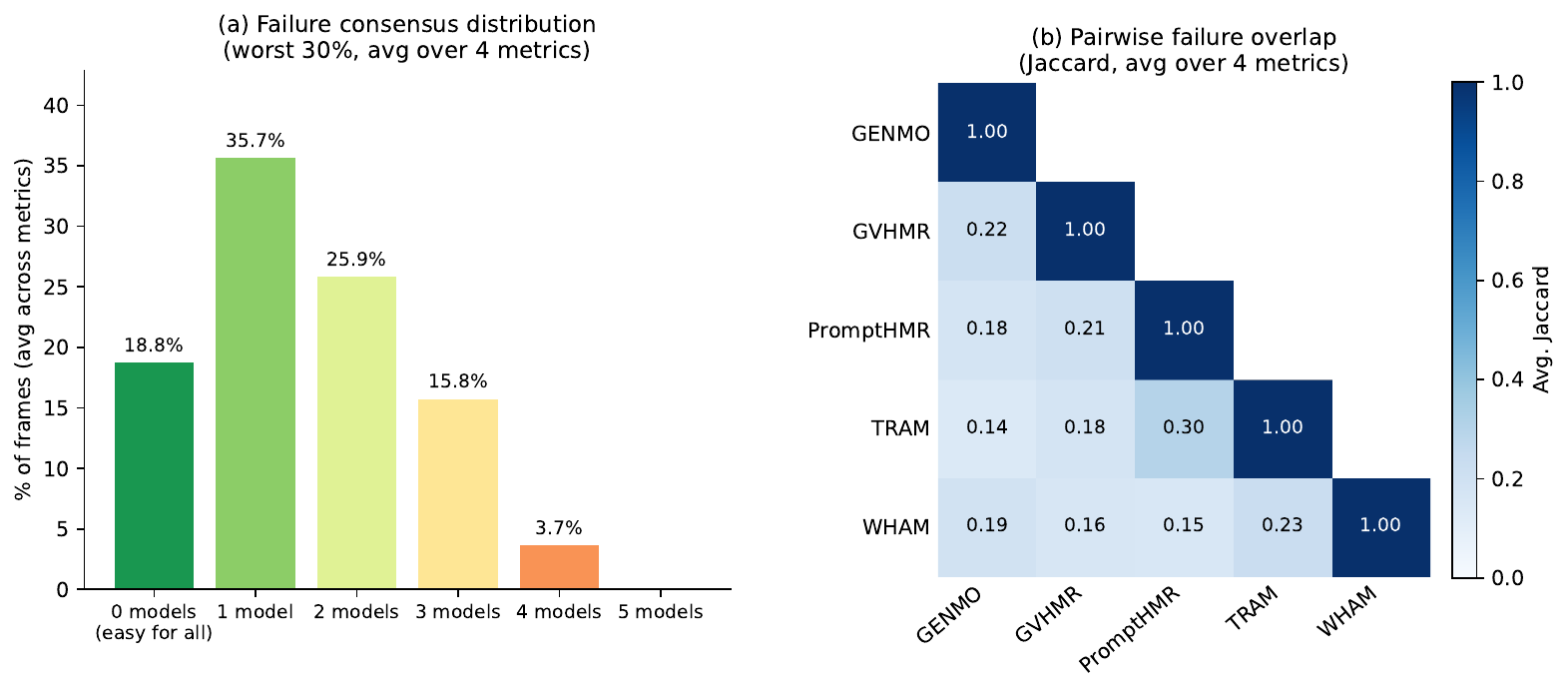}
\caption{Consensus failure distribution (worst 30\% per model, averaged over four metrics).
  \textbf{(a)} $k_f = 0$: frame is in best 70\% for all five models; $k_f = 5$: all five flag it as hard.
  The dominant category ($k_f = 1$, $35.7\%$) and absence of $k_f = 5$ frames confirm that
  failures are model-intrinsic rather than scene-determined.
  \textbf{(b)} Pairwise Jaccard averaged over four metrics confirms consistently low inter-model
  failure correlation.}
\label{fig:consensus}
\end{figure}

Fig.\ref{fig:consensus} shows the consensus failure distribution (averaged
over all four metrics). Since $\tau = 0.30$ is per-model, $k_f = 0$ means a
frame is in the best 70\% for every model simultaneously; $k_f = 5$ would mean
all five simultaneously rank it among their worst 30\%.
$18.8\%$ of frames are uniformly easy ($k_f = 0$), the dominant
category is frames flagged by exactly one model ($k_f = 1$, $35.7\%$), and
no frame reaches $k_f = 5$. The absence of universally hard frames
confirms that failures are model-intrinsic. This confirms our finding in the main analysis that no single model is the best, and model performance varies depending on different scene properties.

\begin{table}[]
\centering
\caption{Notation and Definitions for Failure Signature Analysis}
\label{tab:failure_notation}
\begin{tabular}{ll}
\toprule
\textbf{Variable} & \textbf{Definition} \\
\midrule
$F$              & Set of $N_F$ analysis frames shared across all models \\
$M$              & Set of evaluated models $\{$PromptHMR, WHAM, GVHMR, TRAM, GENMO$\}$ \\
$\varepsilon^m_f$ & Multi-view consistency error of model $m$ at frame $f$ \\
$D^m_f$          & Binary failure label: 1 if $\varepsilon^m_f$ is in the top 30\% for model $m$ \\
$\mathbf{x}_f$   & Scene descriptor vector for frame $f$ (see Table~\ref{tab:scene_features}) \\
$r_s(\mathbf{x}, D^m)$ & Spearman rank correlation between feature $\mathbf{x}$ and failure label $D^m$ \\
$\text{AUC}^m$   & Area under the ROC curve of a logistic classifier using $\mathbf{x}_f$ to predict $D^m_f$ \\
\bottomrule
\end{tabular}
\end{table}

\begin{table}[h]
\centering
\caption{Scene Descriptor Features Extracted per Frame}
\label{tab:scene_features}
\begin{tabular}{lp{9cm}}
\toprule
\textbf{Feature} & \textbf{Description} \\
\midrule
Motion           & Mean optical-flow magnitude across tracked persons \\
Acceleration     & Frame-to-frame change in motion magnitude \\
Depth            & Mean depth of tracked persons from the camera \\
Cam.\ dist.\     & Euclidean distance between the camera center and the mean person position in world coordinates \\
Scale            & Apparent image-plane size proxy (bounding-box height / focal length) \\
Height           & Mean body height of tracked persons in world coordinates \\
P.\ spread       & Spatial spread of predicted body joints in the image plane \\
Foreshortening   & Variance in per-limb foreshortening ratio across the body \\
Blur             & Frame-level sharpness score (Laplacian variance) \\
Mot.\ blur       & Motion-blur magnitude estimated from temporal frame difference \\
\bottomrule
\end{tabular}
\end{table}

\begin{table}[]
\centering
\caption{\textbf{Spearman rank correlations}}
\label{tab:model_comparison}
\vspace{2mm} 
\setlength{\tabcolsep}{0pt} 
\resizebox{\textwidth}{!}{%
\begin{tabular*}{\textwidth}{@{\extracolsep{\fill}} ll rrrrrrrrrrr @{}}
\toprule
\textbf{Model} & \textbf{Metric} & \rotatebox{90}{\small ~Motion} & \rotatebox{90}{\small ~Accel.} & \rotatebox{90}{\small ~Depth} & \rotatebox{90}{\small ~Cam.~dist.} & \rotatebox{90}{\small ~Scale} & \rotatebox{90}{\small ~Height} & \rotatebox{90}{\small ~P.~spread} & \rotatebox{90}{\small ~Foreshortn.} & \rotatebox{90}{\small ~Blur} & \rotatebox{90}{\small ~Mot.~blur} & \textbf{AUC} \\
\midrule
\textbf{PromptHMR} & Trans. & 0.099 & 0.042 & -0.077 & -0.076 & -0.036 & 0.100 & -0.154 & -0.021 & 0.034 & -0.080 & 0.777 \\
 & Pose & -0.001 & -0.003 & -0.062 & -0.083 & 0.159 & 0.005 & 0.153 & -0.051 & 0.032 & \textbf{-0.168} & 0.737 \\
 & PA-M. & 0.022 & -0.024 & -0.122 & -0.159 & 0.199 & 0.013 & 0.006 & -0.011 & 0.019 & -0.142 & 0.690 \\
 & MPJPE & -0.035 & 0.004 & 0.340 & 0.249 & -0.313 & 0.053 & -0.039 & -0.136 & -0.060 & 0.072 & 0.818 \\
\midrule
\textbf{WHAM} & Trans. & -0.005 & 0.011 & 0.252 & 0.218 & -0.115 & 0.077 & 0.125 & 0.013 & 0.123 & -0.089 & 0.761 \\
 & Pose & -0.002 & 0.019 & -0.109 & -0.096 & 0.119 & 0.116 & 0.119 & 0.002 & -0.078 & 0.077 & 0.629 \\
 & PA-M. & -0.009 & 0.011 & 0.021 & 0.017 & -0.024 & 0.038 & -0.037 & 0.076 & -0.030 & 0.023 & 0.657 \\
 & MPJPE & -0.008 & 0.027 & 0.264 & 0.221 & -0.132 & 0.084 & 0.109 & 0.029 & 0.119 & -0.080 & 0.764 \\
\midrule
\textbf{GVHMR} & Trans. & 0.052 & 0.065 & 0.097 & 0.108 & -0.166 & -0.098 & -0.068 & -0.152 & -0.030 & 0.027 & 0.660 \\
 & Pose & -0.022 & -0.056 & -0.155 & -0.164 & 0.218 & 0.049 & 0.003 & -0.005 & 0.073 & -0.068 & 0.700 \\
 & PA-M. & 0.018 & -0.035 & -0.033 & -0.037 & 0.110 & 0.039 & -0.015 & 0.005 & 0.027 & -0.051 & 0.666 \\
 & MPJPE & 0.085 & 0.128 & \textbf{0.463} & 0.381 & \textbf{-0.439} & 0.013 & \textbf{-0.238} & -0.212 & \textbf{0.154} & -0.114 & 0.837 \\
\midrule
\textbf{TRAM} & Trans. & 0.034 & 0.043 & 0.015 & -0.061 & -0.018 & 0.182 & -0.052 & 0.078 & 0.051 & -0.075 & 0.665 \\
 & Pose & 0.044 & 0.019 & -0.066 & -0.056 & -0.017 & -0.106 & -0.026 & -0.070 & -0.050 & -0.037 & 0.673 \\
 & PA-M. & 0.061 & 0.056 & 0.053 & 0.093 & -0.076 & -0.138 & -0.041 & -0.146 & -0.024 & -0.061 & 0.679 \\
 & MPJPE & 0.067 & 0.056 & 0.159 & 0.092 & -0.176 & 0.099 & -0.202 & 0.115 & 0.097 & -0.097 & 0.770 \\
\midrule
\textbf{GENMO} & Trans. & 0.148 & 0.128 & 0.332 & 0.468 & -0.325 & -0.233 & 0.118 & -0.182 & 0.052 & 0.009 & 0.884 \\
 & Pose & 0.090 & 0.059 & -0.050 & 0.059 & 0.187 & -0.174 & 0.114 & 0.060 & -0.054 & 0.034 & 0.773 \\
 & PA-M. & 0.135 & 0.006 & -0.016 & 0.021 & 0.131 & 0.002 & 0.210 & 0.069 & 0.001 & 0.036 & 0.723 \\
 & MPJPE & \textbf{0.166} & \textbf{0.133} & 0.423 & \textbf{0.543} & -0.436 & \textbf{-0.307} & -0.036 & \textbf{-0.260} & 0.091 & -0.071 & \textbf{0.890} \\
\bottomrule
\end{tabular*}%
}
\end{table}

\end{document}